\documentclass[sigplan,twocolumn]{acmart}

\renewcommand\footnotetextcopyrightpermission[1]{}
\usepackage{float}
\usepackage{subfigure}
\usepackage{tikz}
\usepackage{amsmath,amsfonts}
\usepackage{algorithmic}
\usepackage[ruled,linesnumbered]{algorithm2e}
\usepackage{soul}
\usepackage{multirow}
\usepackage{wrapfig}
\usepackage{dutchcal}
\usepackage{pifont}
\usepackage{adjustbox}
\usepackage[normalem]{ulem}

\newcommand{\DEL}[1]{\iffalse #1 \fi}

\newcommand{\sys}{{\scshape FDC}\xspace}

\usepackage{filecontents}

\newcommand{\sh}[1]{\textcolor{blue}{[HS: #1]}}
\newcommand{\red}[1]{\textcolor{red}{#1}}

\usepackage{cleveref}
\crefname{section}{§}{§§}
\Crefname{section}{§}{§§}
\crefformat{section}{§#2#1#3}

\usepackage{ifthenx}

\newcommand*\circled[1]{\tikz[baseline=(char.base)]{
		\node[shape=circle,draw,inner sep=0.5pt] (char) {#1};}}

\newtheorem{obs}{\textbf{Observation}}
\usepackage[most]{tcolorbox}
\definecolor{light-gray}{gray}{0.95}

\usepackage[most]{tcolorbox}
\definecolor{light-gray}{gray}{0.95}

\DEL{
\definecolor{light-gray}{gray}{0.95}
}

\newcommand{\approptoinn}[2]{\mathrel{\vcenter{
  \offinterlineskip\halign{\hfil$##$\cr
    #1\propto\cr\noalign{\kern2pt}#1\sim\cr\noalign{\kern-2pt}}}}}

\usepackage{url}

\newcommand{\squishlist}{
\begin{list}{$\bullet$}
  { \setlength{\itemsep}{0pt}
     \setlength{\parsep}{0pt}
     \setlength{\topsep}{0pt}
     \setlength{\partopsep}{0pt}
     \setlength{\leftmargin}{0em}
     \setlength{\labelwidth}{0em}
     \setlength{\labelsep}{0.2em} } }

\newcommand{\squishlisttwo}{
\begin{list}{$\bullet$}
  { \setlength{\itemsep}{0pt}
     \setlength{\parsep}{0pt}
    \setlength{\topsep}{0pt}
    \setlength{\partopsep}{0pt}
    \setlength{\leftmargin}{2em}
    \setlength{\labelwidth}{1.5em}
    \setlength{\labelsep}{0.5em} } }

\newcommand{\squishend}{
  \end{list}  }

\newlength{\defaultcolumnsep}
\acmConference[Submitted for review to EuroSys]{}
\acmYear{2026}

\begin{document}



\title{\sys: Fast KV Dimensionality Compression for Efficient LLM Inference}

\author{Zeyu Zhang}
\affiliation{
  \institution{University of Virginia}
  \country{USA}
}

\author{Haiying Shen}
\affiliation{
  \institution{University of Virginia}
  \country{USA}
}



\begin{abstract}
\DEL{The prevalent challenge encountered during the inference
phase of large-language models (LLMs) is memory constraints within key-value cache (KVC). Furthermore, the presence of long prompts exacerbates the KVC constraint problem. In this work, we observed that compressing KV values is more
advantageous than compressing model regarding accuracy
and job completion time (JCT). However, we found that KV value quantization and KVC eviction (i.e.,  dropping less-important tokens) generate considerable runtime computational overhead, thus significantly delaying JCT. The methods also cannot handle high network communication time overhead in sequence-parallelism framework for long prompts. To address the problems, in this paper, based on our insightful observations from experimental analysis, we propose  a Zero-delay QKV Compression system for mitigating KV cache and network bottlenecks in LLM inference (\sys). \sys employs the Singular Value Decomposition (SVD) to derive compression matrices, and embeds the compression and decompression operations in model operations. It also adaptively determines the compression rate in the hybrid model layer and token levels in a lightweight manner. Further, it facilitates
creating a communication-efficient sequence parallelism framework for long prompts by reducing communication overhead.  Our trace-driven experiments show that \sys achieves up to 80\% lower JCT and 35\% lower perplexity than state-of-the-art methods and also can complement the methods by reducing their JCT \red{by up to 51\%.}.
}


In large-language models, memory constraints in the Key-Value Cache (KVC) pose a challenge during inference. In this work, we propose \sys, a fast KV dimensionality compression system that eliminates the decompression overhead incurred in the existing KV dimensionality compression system, Palu, and reduces attention time.
Moreover, \sys employs adaptive compression, tailoring KV compression rates across heads and layers based on their contributions to inference to maximize overall compression while maintaining an accuracy loss constraint. Additionally, \sys enhances the attention kernel to balance the uneven workloads caused by the adaptive compression approach to further reduce attention computation latency.
Comprehensive experiments demonstrate that compared to Palu, \sys can reduce Job Completion Time (JCT) by up to 64\%, and delivers up to 1.97$\times$ throughput under the same latency, while maintaining 99\% of the accuracy without compression.
When state-of-the-art eviction and quantization methods are combined with \sys, they exhibit similar improvements compared to those combined with Palu. We open-sourced the code. \looseness=-1

\end{abstract}


\maketitle

\DEL{With a 99\% baseline accuracy constraint, \sys further reduces KV size by up to 68\%, TTFT by up to 44\%, and TBT by up to 55\% compared to the state-of-the-art methods, while delivering up to 1.72$\times$ throughput under the same latency. }



\DEL{Introduction:

It is more important to compress KV values than model. \obsref{obs:qkv_mem_comm_comp}

Not only for memory saving but also for communication saving, and possibly computation saving.

When sequence length is longer, SP is used, the high communication becomes bottleneck. QKV compression can save the communication.

In the introduction, add a fig to show  the effectiveness of our method compared to the most-updated compression approach.

Existing methods for compressing KV values are based on dropping token or quantization.\obsref{obs:time_overhead_qkv_compression}

They have high computation overhead for decision making, compression, and decompression.

When sequence parallelism is used, they cannot save the communication. also show these figs in the introduction
We propose a method to handle this problem

Experimental analysis:
Your current writing is not analysis, usually, you should not put your method there. Analysis is to provide motivation/support, rationale or prove some assumptions to support your design.

Show figs for the memory usage for model and for KV values like vLLM paper - to show it is more important to compress KV values than model.
Not only for memory saving but also for communication saving, and compressing QKV has a  lower computation overhead than compressing models.
Figs to show  using the more advanced approach to compress KV values can have high computation overhead for compression decompression.
Draw figs I asked last night to support your design
Method is combining QK as the first pair for compression and decompression, and combining another pair of two matrix V and L for compression and decompression.
The purpose of combining V and L is to find a way to compress V and decompress V without introducing any compression and decompression overhead.
The time overhead of computing the Y on all vectors and that of computing on part of vectors.
}

\vspace{-0.1in}
\section{Introduction}\vspace{-0in}


Generative large language models (LLMs) \cite{gpt, bert, gpt-2, gpt-3, gpt-4, llama2} have transformed natural language processing (NLP), showcasing remarkable performance across various tasks like text completion \cite{textsynth}, dialogue generation \cite{chatgpt}, code synthesis \cite{github-copilot}, language translation \cite{transformer-translation}, text summarization \cite{egonmwan2019transformer, zi2022source, zhang2022hegel}, and document classification \cite{adhikari2019docbert, dai2022revisiting}. The LLM models are primarily inspired by the transformer architecture \cite{transformer}, which employs attention to capture long-range dependencies in sequences.




During LLM inference, text generation starts with providing a prompt. LLM generates the first token via prompt processing (a.k.a., prefill), followed by text generation (a.k.a., decode) to generate new tokens. LLMs typically consist of multiple transformer layers, where each layer receives an embedding matrix $E$ for the entire input. These embeddings are projected 
to produce Query (Q), Key (K), and Value (V) components for each token, serving as inputs to the attention layer.
The KV values of newly generated tokens are stored in the Key-Value Cache (KVC) to avoid redundant computations. 
During token generation, the KV values of previous tokens are fetched from the KVC and fed into the attention layer. The resulting attention output guides token generation.


\DEL{The embedded matrix for the whole token sequence input is $\mathbf{E}$, where $e_i$ is the embedding of the each token in the sequence.}

The primary challenge of LLM inference is memory constraints within KVC, which limits GPU utilization and consequently decreases throughput. To address it, various KV compression methods have been proposed and they can be divided into two main categories: token eviction methods~\cite{h2o2023zhang, ge2023model, scissorhands2023liu, pyramidinfer, l2norm-kv, keyformer, dynamic-context-pruning, infinigen, zhang2024efficientsparseattentionneeds, jiang2024minference}, quantization methods~\cite{zipcache, kang2024gear, kivi, cachegen, kvquant}, and low-rank approximation methods~\cite{palu}. Token eviction methods remove the KV values of unimportant tokens that have minimal impact on the inference results. Quantization methods compress KV values by reducing the bit width of each element in the KV matrix from a higher bit width (e.g., float16) to a lower bit width (e.g., 4-bit, 2-bit).
However, these methods overlook the redundancy inherent in KV representations~\cite{palu}. By leveraging this redundancy, it is possible to represent KV using fewer hidden dimensions, thereby achieving compression without incurring significant information loss. Recently, Palu~\cite{palu} employs low-rank approximation to compress the hidden dimensions of KV. This approach is complementary to eviction and quantization methods and can be integrated with them to enhance the KV compression rate further~\cite{palu}. InfiniGen~\cite{infinigen} uses low-rank approximation to analyze attention scores and identify unimportant tokens efficiently; however, it is fundamentally an eviction method.
\looseness=-1

Currently, state-of-the-art LLMs adopt Rotary Positional Embedding (RoPE)~\cite{roformer} to inject positional information into tokens' $Q$ and $K$. However, as explained in~\cref{sec:bkg}, the presence of RoPE necessitates that Palu first decompress the compressed $K$ before RoPE, and only then can the RoPE-processed $K$ be fed into the attention. This leads to two main issues: 1) an unavoidable $K$-decompression overhead, and 2) the inability to directly use the compressed $K$ to accelerate attention~\cite{palu}. Because many LLM tasks, such as summarization, book generation, and Information Retrieval (IR), involve long-sequence processing~\cite{arxiv-summarization, cocktailforir, pg-19}, the context window length of state-of-the-art LLMs now extends to 128K or even 1M tokens~\cite{mistral-v0.3, llama3.1, phi-3, yi-model, gemmateam2024gemma2improvingopen}. Long sequences amplify Palu's $K$-decompression overhead and the attention overhead. Experimental results in~\cref{sec:palu_overhead} show that for input lengths up to 100K, Palu’s attention computation can account for up to 36.3\% and 31.9\% of the Job Completion Time (JCT) during the prefill and decode phases, respectively, and decompressing $K$ during the decode phase can reach up to 19.7\% of the JCT. For output lengths up to 100K, Palu’s attention computation in the decode phase can climb to 74.6\% of the JCT, while the time to decompress $K$ can reach up to 29.5\%.

We propose \sys to tackle these issues. \sys is a \uline{F}ast KV \uline{D}imensionality \uline{C}ompression system, which not only eliminates the $K$-decompression overhead but also reduces attention time for KV dimensionality compression. \sys comprises three key components as below.

\textbf{Post-RoPE QK compression for decompression elimination and attention acceleration.}
Unlike Palu, which compresses QK before RoPE, \sys compresses QK after RoPE. In~\cref{sec:opportunities}, we observe that the post-RoPE QK is also compressible, which is determined by the model itself rather than the input tokens. As a result, a rotation matrix $R$ suitable for compressing post-RoPE QK can be precomputed offline. During online inference, the post-RoPE QK is compressed using $R$, and the compressed QK is fed directly into attention, eliminating the decompression of $K$. Attention is computed on the compressed QK, reducing its computation time. \looseness=-1


\textbf{Adaptive compression rate.} Model layers contribute differently to the output~\cite{transformer}, as do the attention heads within each layer~\cite{ge2023model}. Thus, \sys applies different compression rates to QKV in different heads and layers to maximize the overall KV compression rate given an accuracy constraint. These compression rates are determined by the magnitude of the components in each dimension of the compressed QKV vectors.

\textbf{Attention kernel enhancement for adaptive compression rate.} The different compression rates on the hidden dimensions of QKV in \sys can lead to imbalanced execution times across a GPU's Stream-Multiprocessors (SM). Thus, \sys balances the uneven workloads across SMs for attention in the prefill and decode phases, respectively, to further reduce attention computation latency.

\DEL{Further building on our findings, which indicate that: 1)  The experiment shows that our method can have up to ??\% lower JCT and ??\% lower perplexity increase than the existing methods. }

In summary, our work has the following contributions:
\squishlist
\item We conduct extensive experiments with numerous state-of-the-art models and complex-task datasets to demonstrate the rationality and feasibility of \sys's design principles and provide geometric explanations for these observations.

\item We propose \sys, a fast KV dimensionality compression system that eliminates the $K$-decompression overhead and accelerates attention. It can complement existing token eviction based and quantization based KV compression methods.


\item Comprehensive experiments demonstrate that compared to Palu, \sys can reduce JCT by up to 64\%, Time-To-First-Token (TTFT) by up to 35\%, and Time-Between-Token (TBT) by up to 64\%, and delivers up to 1.97$\times$ throughput under the same latency, while maintaining 99\% of the accuracy without compression.
When the state-of-the-art eviction and quantization methods are combined with \sys, they exhibit similar improvements compared to those combined with Palu.

\squishend

We open-sourced the code of \sys~\cite{zeroc-code}.

\section{Background}\label{sec:bkg}


We explain how the current work compresses KV along the hidden dimension. Table~\ref{tab:symbols} lists notations used in the paper.

\begin{table}[h]
\centering
\small
\begin{tabular}{ |c|l||l|l|  }
 \hline
 $d$ & Model dimension size & $N_h$ & The number of heads\\
 \hline
 $N_l$ & The number of layers & $d_h$ & Head dimension size ($=d/N_h$) \\
 \hline
 $E$ & Token embeddings & $Q$ & Query matrix \\
 \hline
 $K$ & Key matrix & $V$ & Value matrix \\
 \hline
 $R$& Rotation matrix& $p$& Compression rate\\
 \hline
 $W$& Parameter matrix& $s$& Sequence length \\
 \hline
 \end{tabular}
 \vspace{-0in}
\caption{Notations used in the paper.}\label{tab:symbols}
\vspace{-0.15in}
\end{table}

The common way to compress the hidden dimension of KV is first applying Singular Value Decomposition (SVD)~\cite{svd} to the parameter matrices that generate KV~\cite{palu}. Because the procedures for $K$ and $V$ are identical, we describe the process for $K$ only. Let $K=EW$, where $W$ is the parameter matrix for $K$. The shapes of $E$ and $W$ are $s\times d$ and $d\times d_h$, respectively. Decomposing $W$ via SVD yields two matrices $A$ and $B$ such that $W=AB$, with shapes $d\times k$ and $k\times d_h$, where $k<d_h$. The original $K$ has shape $s\times d_h$. During compression, the compressed $K'$ is generated via $K'=EA$, whose shape is $s\times k$. Because $k<d_h$, $K'$ is smaller than $K$, achieving the desired compression. To reconstruct an approximation of the original $K$, $K'$ and $B$ are multiplied, i.e., $K'B\approx K$. Palu~\cite{palu} is the state-of-the-art and representative method that employs this form of KV compression.

Currently, the state-of-the-art transformer-based LLMs employ RoPE~\cite{roformer} to encode positional information for tokens. RoPE processes the generated $Q$ and $K$. Specifically, RoPE rotates each row vector \textbf{\textit{q}} and \textbf{\textit{k}} of $Q$ and $K$ using a rotation matrix $H_i$, where $i$ is the position of \textbf{\textit{q}} and \textbf{\textit{k}} in the sequence.
Without RoPE, the computation $QK^T$ in attention can be rewritten as $EW_Q(EW_K)^T=EW_Q(EA_KB_K)^T=E(W_QB_K^T)K'^T$, where $W_Q$ and $B_K$ are merged together offline to remove the online decompression overhead for multiplying $B_K$~\cite{palu}. And $K'$ can be used directly in attention to reduce the computation time. However, the existence of RoPE prevents the integration of $B_K$ into $W_Q$ because $QK^T$ in attention is changed to $\mbox{RoPE}(Q)\mbox{RoPE}(K)^T$~\cite{palu}. Therefore, Palu keeps the decompression process $K'B_K$ before RoPE, which introduces the decompression overhead for $K'$. Besides this, Palu's attention takes $\mbox{RoPE}(Q)$ and $\mbox{RoPE}(K)$ as inputs for computation. Hence, the attention cannot leverage the compressed $K'$ for acceleration. Long sequences can amplify this decompression overhead and attention time. We explain it in \cref{sec:motivation}.

\section{Motivation}\label{sec:motivation}


In this section, we analyze Palu’s $K$-decompression overhead as well as the overhead associated with attention computation for RoPE-based models, and we highlight potential opportunities for optimizing Palu.

\subsection{Decompression and Attention Overhead of Palu}\label{sec:palu_overhead}

\begin{figure}[h]\vspace{-0.05in}
    \centering
    \subfigure[Output length 256 with varying input lengths.\label{fig:diff_input_length}]
    {\includegraphics[width=0.495\columnwidth,height=2.4cm]{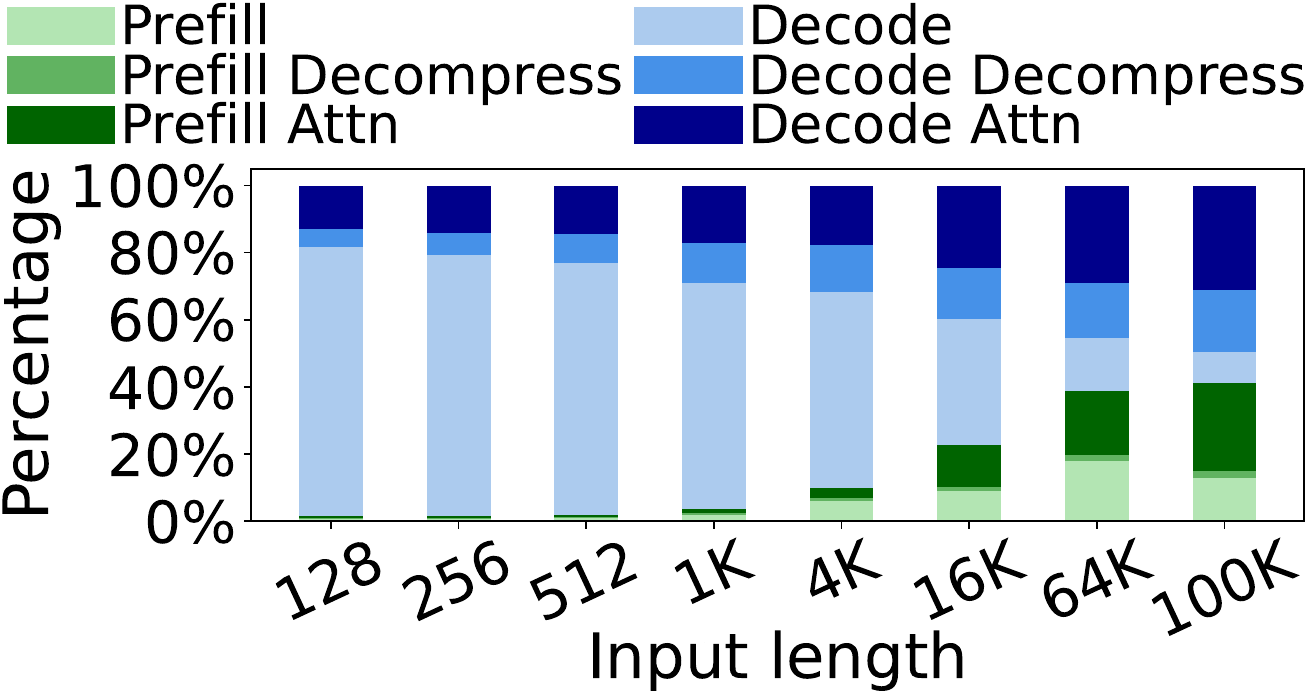}}
    \subfigure[Input length 256 with varying output lengths.\label{fig:diff_output_length}]
    {\includegraphics[width=0.495\columnwidth,height=2.4cm]{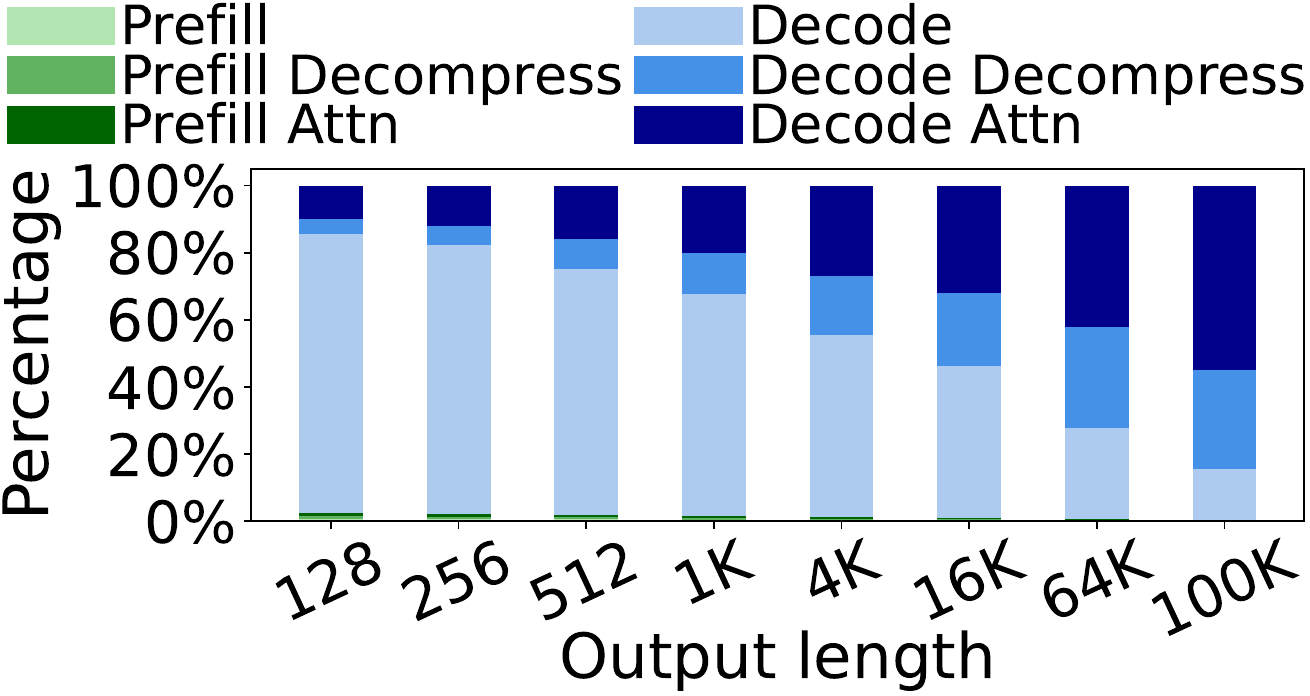}}

    \vspace{-0.15in}
    \caption{Palu's JCT decomposition for Llama-3.1 70B with different input and output lengths.}
    \label{fig:diff_input_output_length} \vspace{-0.2in}
\end{figure}

We first selected the largest model with 128K context window length, Llama-3.1 70B, from Table~\ref{tab:models} to examine the percentage of each JCT component under different input and output lengths. The ranges for input and output lengths were determined based on the dataset information in Table~\ref{tab:datasets}. When varying input and output lengths, the corresponding output length and input length were set to 256, which are the common lengths in datasets in Table~\ref{tab:datasets}. Fig.~\ref{fig:diff_input_output_length} illustrates the JCT decomposition of Palu under various input and output length settings for Llama-3.1 70B.

Fig~\ref{fig:diff_input_length} shows that as the input length increases from 128 to 100K, the time spent in decompressing $K$ during the decode phase grows from 5.6\% to 19.7\% of the JCT. The percentage of attention time during the prefill phase rises from 0.4\% to 26.3\% of the JCT, and the percentage of attention time during the decode phase increases from 13.1\% to 29.5\%. This is because, with longer input lengths, the overhead of decompressing $K$ in the decode phase grows, and the attention computation overhead during both prefill and decode phases also increases. The decompression time during the prefill phase accounts for no more than 2\% of the JCT, mainly because the prefill phase has only one iteration (and thus decompresses $K$ only once), unlike the decode phase that involves multiple iterations.

Fig.~\ref{fig:diff_output_length} shows that as the output length increases from 128 to 100K, the proportion of attention computation during decode grows from 11.3\% to 55.7\% of the JCT, while the time spent in decompressing $K$ increases from 4.4\% to 29.5\% of the JCT. This occurs because a longer output length leads to larger $K$, increasing both the attention computation time and the time required to decompress $K$. The prefill time remains under 2\% of the JCT, which is due to the large number of decode iterations (associated with a long output), causing the prefill phase to account for only a small fraction of JCT. Therefore, in the case of long output lengths, the main focus should be on improving the decode phase.

\begin{figure}[h]\vspace{-0.05in}
    \centering
    \subfigure[Input length 100K and output length 256.\label{fig:diff_model_long_input}]
    {\includegraphics[width=0.495\columnwidth,height=2.4cm]{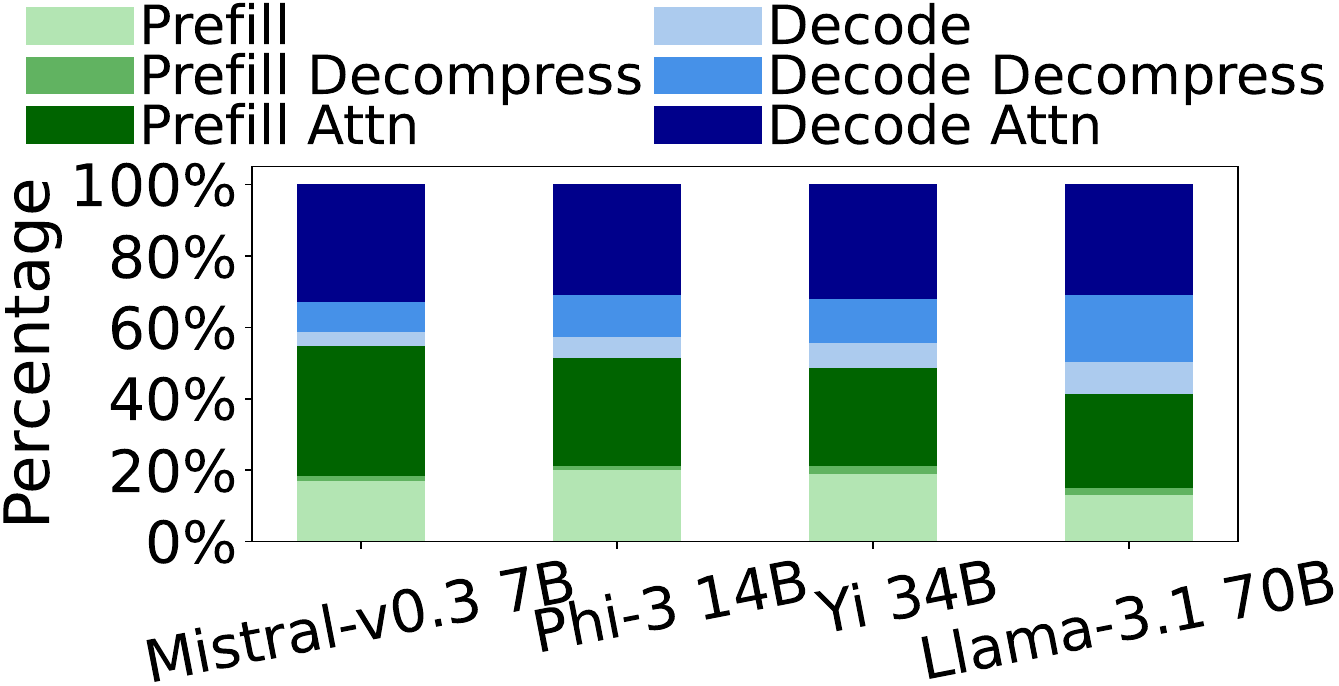}}
    \subfigure[Input length 256 and output length 100K.\label{fig:diff_model_long_output}]
    {\includegraphics[width=0.495\columnwidth,height=2.4cm]{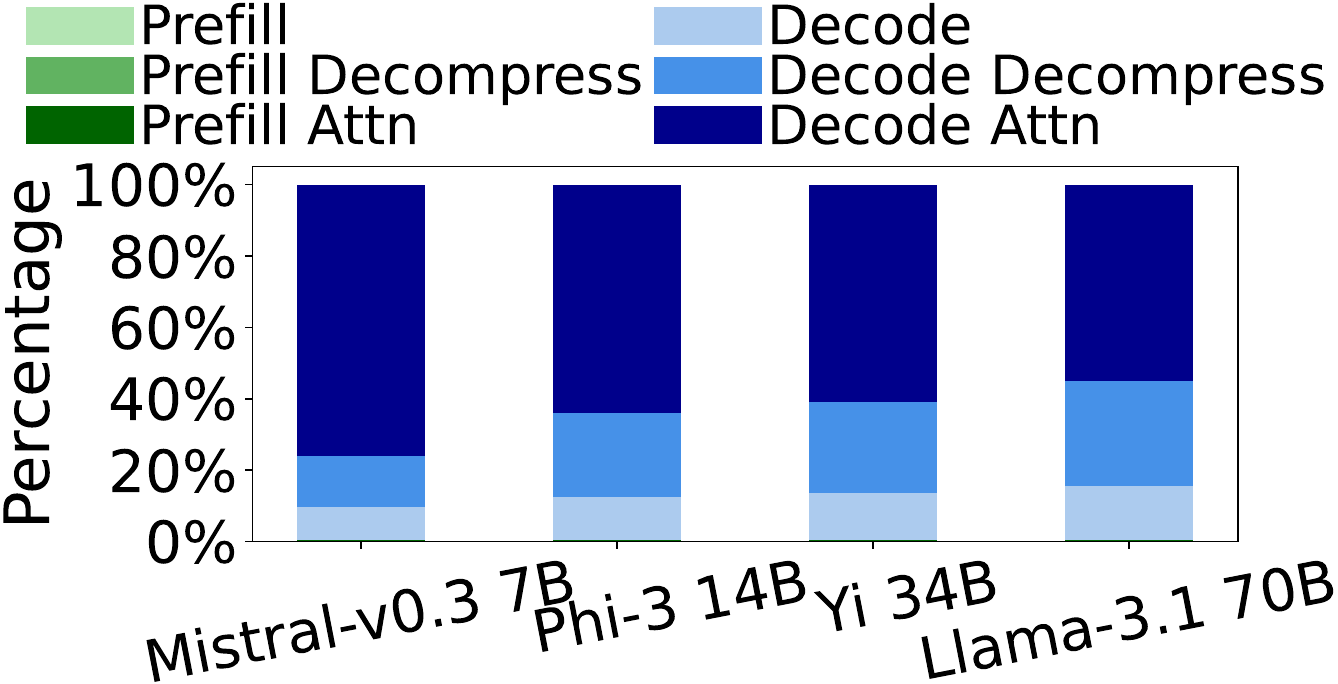}}

    \vspace{-0.15in}
    \caption{Palu's JCT decomposition for different models.}
    \label{fig:diff_model_long_length} \vspace{-0.1in}
\end{figure}

Additionally, we employed various models to investigate how a long input length and a long output length influence the components of JCT. The models used are those in Table~\ref{tab:models} that support a context window length of 128K. Fig.~\ref{fig:diff_model_long_input} illustrates the scenario where the input length is 100K and the output length is 256. As the model size increases from 7B to 70B, the attention computation time during the prefill and decode phases drops from 36.3\% and 31.9\% of JCT, respectively, to 26.3\% and 29.5\%. This is because a larger model has a larger hidden dimension $d$, which causes the linear operations that scale quadratically with $d$ to occupy a greater portion of the total time (whereas attention scales linearly with $d$). Meanwhile, the time spent in decompressing $K$ during decode increases from 10.5\% to 19.7\% of the JCT, as a larger $d$ leads to a larger-sized $K$ and thus more decompression time. During the prefill phase, the decompression time remains under 2.2\% of the JCT, for the same reason described when discussing Fig.~\ref{fig:diff_input_length}.

Fig.~\ref{fig:diff_model_long_output} illustrates the case where the input length is 256 and the output length is 100K. As the model size increases from 7B to 70B, the attention time during decode decreases from 74.6\% to 55.7\% of the JCT, while the decompression time increases from 14.4\% to 29.5\%. This is the same reason as given for Fig.~\ref{fig:diff_model_long_input}.

Overall, long sequences amplify both Palu’s attention overhead and the overhead associated with decompressing $K$. For input lengths up to 100K, Palu’s attention computation can account for up to 36.3\% and 31.9\% of the JCT during the prefill and decode phases, respectively, and decompressing $K$ during the decode phase can reach up to 19.7\% of the JCT. For output lengths up to 100K, Palu’s attention computation in the decode phase can climb to 74.6\% of the JCT, while the time to decompress $K$ can reach up to 29.5\%.

\subsection{Opportunities}\label{sec:opportunities}

To reduce the attention overhead, it is desirable to perform computation directly using the compressed $K$. Doing so also avoids the computational cost of decompressing K prior to the attention. Therefore, we may consider compressing $K$ after it has been processed by RoPE. However, this approach should not introduce additional overhead at the end of the attention computation; otherwise, the benefits gained from removing $K$ decompression would be diminished.

Let us denote the post-RoPE $Q$ and $K$ as $Q_p$ and $K_p$, respectively. Suppose there is a rotation matrix $R$ with shape $d_h \times d_h$. Then, the attention computation $Q_pK_p^T$ is mathematically equivalent to $(Q_pR)(K_pR)^T$ since $(Q_pR)(K_pR)^T=Q_p(RR^T)K_p^T=Q_pK_p^T$. In other words, computing $Q_pK_p^T$ can be reformulated as multiplying $Q_pR$ and $K_pR$.
If the matrix $R$ rotates all row vectors of $Q_p$ and $K_p$ such that the resulting vectors contain many low-magnitude dimensions, those dimensions can be removed to compress $Q_p$ and $K_p$ without much information loss. This allows us to compute an approximate value of $Q_pK_p^T$ directly by multiplying the compressed versions of $Q_p$ and $K_p$. This approach will be detailed in~\cref{sec:design}.

However, achieving this objective requires satisfying two conditions. First, $Q_p$ and $K_p$ must be compressible. That is, there must exist a common rotation matrix $R$ such that the rotated versions ($Q_pR$ and $K_pR$) contain many low-magnitude dimensions whose removal does not significantly distort the data. Second, the common $R$ used for $Q_p$ and $K_p$ should be model-specific and independent of the tokens. In other words, given a model, this common $R$ can be determined offline and will not vary with the dataset.
These two conditions raise two critical questions.
\squishlist
\vspace{-0.0in} \item[1)] Does there exist a common $R$ that can compress $Q_p$ and $K_p$ without significantly distorting the data?

\vspace{-0.0in} \item[2)] Is the common $R$ that $Q_p$ and $K_p$ share model-specific and independent of the tokens?

\squishend
The answer to both questions is yes. We will address them in the following observations, which give us opportunities for optimizing Palu.

\noindent \textbf{Does there exist a common $R$ that can compress $Q_p$ and $K_p$ without significantly distorting the data?}
We observe that the answer is yes. In the NLP domain, it is common to apply SVD directly to a matrix $X$ composed of row vectors (e.g., word embeddings). This decomposition yields singular values and an $R$, which can then be used to compress $X$. This approach is detailed in~\cref{sec:svd_for_compression}.
The magnitudes of the singular values reflect the magnitude of the row vectors of $XR$ along each dimension. If many singular values approach zero, it indicates that $XR$ contains a large number of low-magnitude dimensions that can be discarded without significantly distorting the data.

\begin{figure}[h]\vspace{-0in}
  \centering
  \includegraphics[width=0.9\columnwidth]{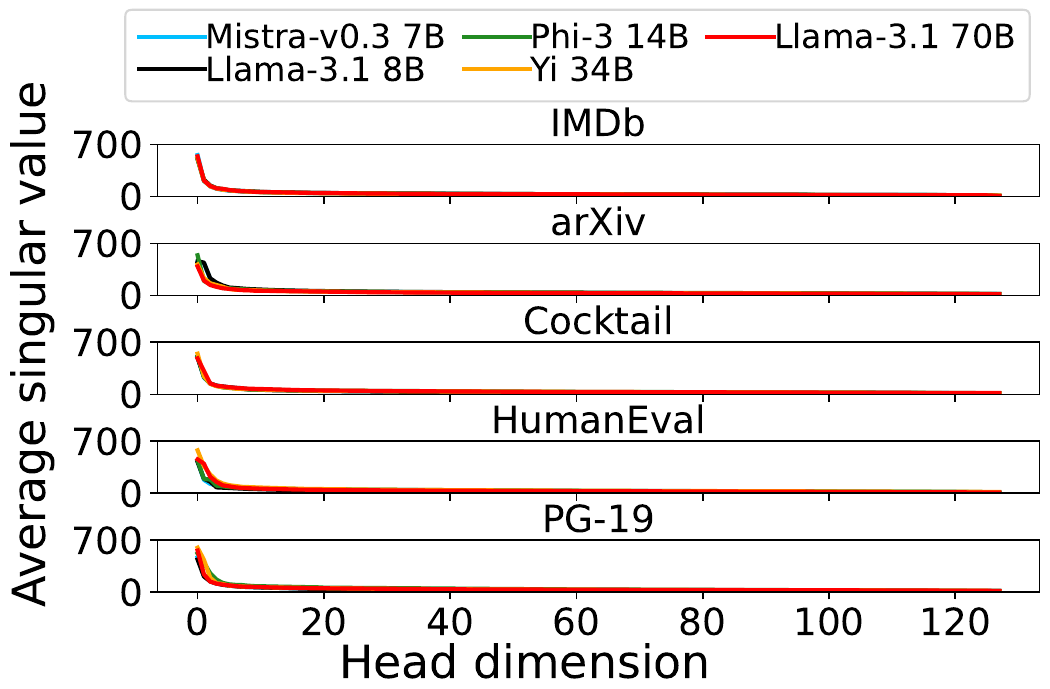}
  \vspace{-0.1in}
  \caption{The average singular value for the $Q_p$-$K_p$ pair.}
  \label{fig:singular_value_for_qk}
  \vspace{-0.1in}
\end{figure}

Following this approach, we combine the row vectors of $Q_p$ and $K_p$ into a single matrix and calculate its singular values and $R$ for each attention head.
Fig~\ref{fig:singular_value_for_qk} shows the average singular values across all attention heads for different models in Table~\ref{tab:models} and datasets in Table~\ref{tab:datasets}.
We observe that the average singular values in the leftmost dimensions are significantly higher than those in other dimensions, indicating that the row vectors of $Q_pR$ and $K_pR$ have many small-magnitude dimensions whose removal does not change the vectors much.
The explanation for this observation will be detailed in~\cref{sec:q1}.

\noindent\textbf{Is the common $R$ that $Q_p$ and $K_p$ share model-specific and independent of the tokens?}
We observe that the answer is yes.
To investigate this question, we need to examine whether the matrices $R$ generated from different datasets are similar. If $R$ is independent of the token, then the $R$ derived from different datasets should exhibit a certain degree of similarity. Moreover, the $R$ computed based on randomly generated tokens should also resemble these $R$ derived from datasets. To validate this, we randomly generate a sequence of tokens from the vocabulary, with the total number of tokens matching that of a randomly selected dataset. Based on these random tokens, we compute a rotation matrix $R_r$. Each dataset yields its own $R$. We then measure the degree of difference between each dataset-specific $R$ and $R_r$. This difference is quantified by the ratio of the deviation in each element of $R$ to its magnitude, where smaller ratios indicate lower degrees of difference.

Specifically, given a model, each head $h$ in layer $l$ has an $R^{h,l}_r$ generated from random tokens and an $R^{h,l}_a$ from a dataset $a$.
We use AABS($M$) to denote the average absolute value of all elements in matrix $M$ and SUB($M_1$,$M_2$) to denote the element-wise subtraction of $M_2$ from $M_1$. For a model and a dataset $a$, we compute $\overline{\varepsilon}=\frac{1}{N_hN_l}\sum_{h=1}^{N_h}\sum_{l=1}^{N_l}$AABS($R^{h,l}_a$) to evaluate the average magnitude of an individual element in $R^{h,l}_a$.
We also calculate $\overline{\delta}=\frac{1}{N_hN_l}\sum_{h=1}^{N_h}\sum_{l=1}^{N_l}$AABS(SUB($R^{h,l}_{r}$,$R^{h,l}_{a}$)) to evaluate the average magnitude of the differences between the elements of $R^{h,l}_{r}$ and $R^{h,l}_{a}$.

\begin{table}[h]\vspace{-0in}
\centering
\renewcommand{\arraystretch}{1.8}
\begin{adjustbox}{max width=\columnwidth}
\begin{tabular}{|c||c|c|c|c|c|}
\hline
$\overline{\delta}/\overline{\varepsilon}$ & IMDb & arXiv & Cocktail & HumanEval & PG-19 \\
\hline
Mistral-v0.3 7B & $\displaystyle\frac{8.3e-5}{6.43e-2}$ & $\displaystyle\frac{1.8e-4}{6.41e-2}$ & $\displaystyle\frac{7.1e-5}{6.44e-2}$ & $\displaystyle\frac{4.5e-5}{6.43e-2}$ & $\displaystyle\frac{5.6e-5}{6.39e-2}$ \\
\hline
Llama-3.1 8B & $\displaystyle\frac{1.1e-4}{5.29e-2}$ & $\displaystyle\frac{1.5e-4}{5.27e-2}$ & $\displaystyle\frac{8e-5}{5.28e-2}$ & $\displaystyle\frac{1.7e-4}{5.3e-2}$ & $\displaystyle\frac{1.4e-4}{5.25e-2}$ \\
\hline
Phi-3 14B & $\displaystyle\frac{1.9e-4}{5.41e-2}$ & $\displaystyle\frac{5.2e-5}{5.43e-2}$ & $\displaystyle\frac{1.1e-4}{5.44e-2}$ & $\displaystyle\frac{4.6e-5}{5.44e-2}$ & $\displaystyle\frac{1.3e-4}{5.41e-2}$ \\
\hline
Yi 34B & $\displaystyle\frac{1.7e-4}{5.05e-2}$ & $\displaystyle\frac{4.9e-5}{5.08e-2}$ & $\displaystyle\frac{7.6e-5}{5.06e-2}$ & $\displaystyle\frac{1.2e-4}{5.06e-2}$ & $\displaystyle\frac{8.9e-5}{5.11e-2}$ \\
\hline
Llama-3.1 70B & $\displaystyle\frac{9.5e-5}{5.54e-2}$ & $\displaystyle\frac{1.3e-4}{5.53e-2}$ & $\displaystyle\frac{8.4e-5}{5.54e-2}$ & $\displaystyle\frac{1.6e-4}{5.52e-2}$ & $\displaystyle\frac{1.02e-4}{5.51e-2}$ \\
\hline
\end{tabular}
\end{adjustbox}
\vspace{-0in}
\caption{$\overline{\varepsilon}$ and $\overline{\delta}$ for the $Q_p$-$K_p$ pair.}
\vspace{-0.2in}
\label{tab:r_diff_qk}
\end{table}

Table~\ref{tab:r_diff_qk} presents $\overline{\varepsilon}$ and $\overline{\delta}$, displayed in the format $\overline{\delta}/\overline{\varepsilon}$. We observe that the average variation $\overline{\delta}$ in the values of individual elements of $R$ is less than 0.34\% of the average absolute value $\overline{\varepsilon}$ of a single element. This upper limit is derived from the results for Yi 34B with IMDb. This indicates that the $R_p$ derived from random tokens is highly similar to the $R$ derived from a different dataset, with minimal variation. Therefore, the $R$ is independent of tokens. In other words, the $R_r$ computed offline from random tokens can be applied to other datasets to compress $Q_p$ and $K_p$ online. The explanation for this observation is detailed in~\cref{sec:q2}.

\noindent
To sum up, post-RoPE $Q_p$ and $K_p$ have a common rotation matrix $R$ that can compress them without significantly distorting the data. The common $R$ is independent of the tokens and only specific to the model. These two observations give us opportunities to accelerate attention and eliminate the $K$-decompression overhead in Palu.

\vspace{-0in}
\section{Understanding Observations}\label{sec:understand_obs}
\vspace{-0.03in}


In this section, we provide essential information and explanations to facilitate a deeper understanding of the observations presented in the motivation.

\vspace{-0in}
\subsection{SVD for Data Compression in NLP}\label{sec:svd_for_compression}
\vspace{-0.03in}


\begin{wrapfigure}[5]{c}{0.32\textwidth}\vspace{-0.15in}
  \centering
  \includegraphics[width=0.32\textwidth,height=1.5cm]{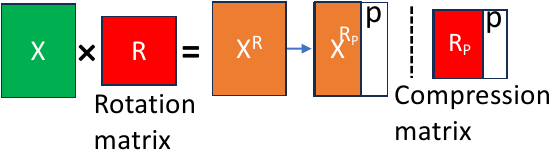}
  \vspace{-0.33in}
  \caption{SVD for data compression.}
  \label{fig:ar_matmul}
\end{wrapfigure}

\setlength{\columnsep}{8pt}
In NLP, SVD is widely used to reduce data dimensionality with low or even zero loss. Given a matrix $X$ where each row represents an item's vector to be compressed, $X$ can be decomposed into $X=U\Sigma R^T$. $U$ and $R$ are rotation matrices, while $\Sigma$ is a diagonal matrix filled with zeros except for non-negative singular values sorted in descending order along the diagonal. This decomposition, solvable using an SVD solver \cite{svd-solver}, has a unique solution. Singular values quantify the magnitude of vectors' components after rotation by $R$. Consequently, each row of $X^R=XR$ is a rotated vector with elements decreasing from left to right, and the right $p$ fraction of the columns of $X^R$ can be discarded with minimal impact on vector relationships. This creates a compressed matrix $X^{R_p}$, with $R_p$ as the compression matrix, as depicted in Fig.~\ref{fig:ar_matmul}. To decompress the data, we only need to use the inverse of $R$, equivalent to $R^T$ for rotation matrices, to rotate the compressed vectors back to their original orientation. Since the deleted dimensions can be considered zero-valued, we do not require the full $R^T$ for decompression; instead, $R_p^T$ suffices. \looseness=-1
\setlength{\columnsep}{\defaultcolumnsep}

For any set of vectors we want to compress, such as all the row vectors of $Q_p$ and $K_p$, we can arrange these vectors as rows to form the matrix $X$ shown in Fig.~\ref{fig:ar_matmul}. We then obtain the singular values and the matrix $R$ used for compressing the vectors following the method described above. As long as the singular values contain many small values, it indicates that the set of vectors can be compressed using $R$ without incurring significant loss.

\subsection{Why Are Post-RoPE $Q_p$ and $K_p$ Compressible?}\label{sec:q1}

In~\cref{sec:opportunities}, our experiments show that the post-RoPE $Q_p$ and $K_p$ exhibit many low-magnitude singular values after SVD decomposition, indicating that $Q_p$ and $K_p$ can be compressed using a matrix $R$ without introducing significant loss. But why do $Q_p$ and $K_p$ exhibit this property? Prior work~\cite{infinigen,palu} has demonstrated that the pre-RoPE $Q$ and $K$ possess a low-rank property, meaning that their SVD decompositions yield many low-magnitude singular values. This suggests that $Q$ and $K$ have many dimensions in the space where their magnitudes are very small.

From a geometric perspective, the RoPE operation applied to $Q$ and $K$ rotates each vector within the plane formed by every pair of dimensions. Such rotational transformations do not amplify the vector magnitudes along the two dimensions that define each plane. Therefore, if a vector initially exhibits low magnitudes along the dimensions forming a plane, the RoPE rotation will preserve these low magnitudes post-transformation. Since $Q$ and $K$ naturally possess many dimensions with low magnitude, after RoPE, the corresponding dimensions with low magnitude that can form a plane in RoPE will also retain low magnitudes. This geometric behavior explains why $Q_p$ and $K_p$ still have low-magnitude dimensions and why the singular values of $Q_p$ and $K_p$ continue to exhibit many small values.

Because $Q_p$ and $K_p$ both have many low-magnitude dimensions, if they share many common low-magnitude dimensions, then performing SVD decomposition on all vectors of $Q_p$ and $K_p$ together will produce singular values with many small values. Fig.~\ref{fig:singular_value_for_qk} has already shown that these singular values contain many small values, indicating that $Q_p$ and $K_p$ have common low-magnitude dimensions and therefore can be compressed using a common $R$.

\begin{figure*}[h]\vspace{-0in}
  \centering
  \includegraphics[width=0.99\textwidth]{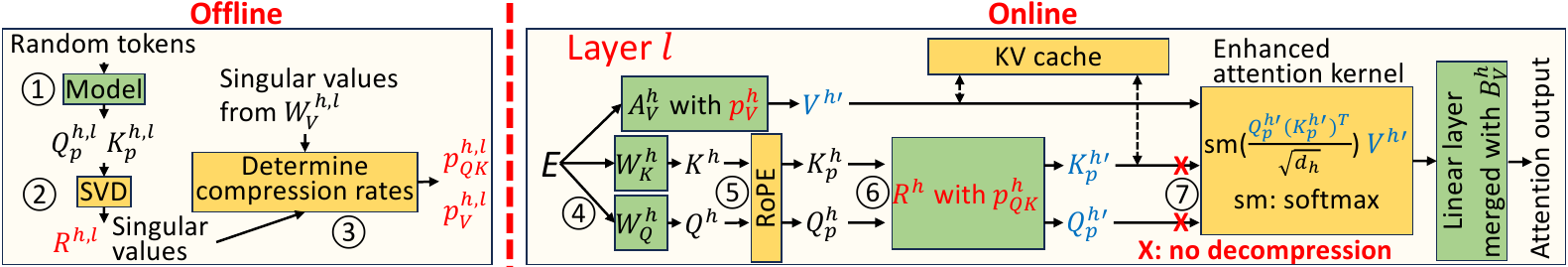}
  \vspace{-0.15in}
  \caption{Overview.}
  \label{fig:design_overview}
  \vspace{-0.1in}
\end{figure*}

\subsection{Why Is the Common $R$ for $Q_p$ and $K_p$ Independent of Tokens?}\label{sec:q2}
\vspace{-0.03in}

The singular values and the common $R$ obtained by performing SVD decomposition on all vectors of $Q_p$ and $K_p$ are correlated. The singular values represent the magnitudes of all dimensions of the vectors after rotation by $R$. To answer why the common $R$ is independent of tokens is to answer why the distribution of singular values is independent of tokens. Since the singular values reflect the dimension magnitudes of $Q_p$ and $K_p$, this question can be further rephrased as why the low-magnitude dimensions of $Q_p$ and $K_p$ are independent of tokens.

Existing work has shown that the pre-RoPE $Q$ and $K$ have a low-rank property, and that the low-magnitude dimensions of $Q$ and $K$ are independent of tokens but related to model parameters~\cite{infinigen, palu}. We take $Q$ as an example to explain the reason. Since $Q = EW$, we can perform SVD decomposition on $W$ to obtain $W = U\Sigma R^T$, resulting in $Q = EU\Sigma R^T$. Here, $\Sigma$ and $R$ are the singular values and the rotation matrix used for compression of $Q$. $\Sigma$ originates from the parameter $W$ and is unrelated to the token embedding $E$, so the distribution of dimension magnitudes in $Q$ depends only on the model parameters and is independent of the tokens. The same holds for K.
Importantly, since the RoPE operation is also independent of the token embedding $E$, the distribution of dimension magnitudes in the post-RoPE $Q_p$ and $K_p$ also depends only on the parameters and is independent of the tokens. Therefore, the singular values and $R$ obtained by decomposing $Q_p$ and $K_p$ are independent of tokens. This also explains why the $R$ obtained using different datasets (even random tokens) is highly similar.

\section{Design}\label{sec:design}


We detail our design in this section. Fig.~\ref{fig:design_overview} gives an overview of our approach, which consists of two stages: an offline stage and an online stage. The offline stage is responsible for obtaining the rotation matrix $R$ used to compress $Q_p$ and $K_p$, as well as the compression rate $p$ for $Q_p$, $K_p$, and $V$ under different accuracy constraints. The online stage performs inference using the $R$ and $p$ obtained during the offline stage under an accuracy constraint.

In the offline stage, we first randomly generate tokens and feed them into the model to extract $Q_p^{h,l}$ and $K_p^{h,l}$ from each head $h$ in layer $l$ (step \circled{1}). We then apply SVD to $Q_p^{h,l}$ and $K_p^{h,l}$ to obtain their singular values and the rotation matrix $R^{h,l}$ (step \circled{2}), as explained in~\cref{sec:design:decompress_attn}. Since different heads contribute unequally to inference results~\cite{ge2023model, transformer}, we determine the compression rate $p$ for $Q_p$, $K_p$, and $V$ based on the singular values (step \circled{3}), with the goal of maximizing the overall KV compression rate under an accuracy constraint. Details are provided in~\cref{sec:Hybrid}. In the online stage, for each layer $l$, token embeddings $E$ are linearly transformed by parameters $W_Q^h$ and $W_K^h$ to produce $Q^h$ and $K^h$ (step \circled{4}), which are then processed by RoPE to generate $Q^h_p$ and $K^h_p$ (step \circled{5}). These are subsequently compressed into $Q_p^{h}{'}$ and $K_p^{h}{'}$ using the rotation matrix $R^h$ and compression rate $p_{QK}^h$ shared by $Q$ and $K$ (step \circled{6}), following the method shown in Fig.~\ref{fig:ar_matmul}. The compressed $Q_p^{h}{'}$ and $K_p^{h}{'}$ are directly used in the attention without requiring decompression.
The compression method for $V$ follows Palu but uses the compression rate $p_V^h$ determined in step \circled{3} to maximize overall KV compression rate. The compressed $V^{h}{'}$ is also fed directly into attention. Since $Q_p^h$, $K_p^h$, and $V^h$ have different compression rates, the workload across SMs of the GPU during attention computation may become imbalanced. To address this, we design an enhanced attention kernel that balances SM workloads to accelerate attention computation (step \circled{7}). Details are provided in~\cref{sec:kernel_opt}.

\subsection{Post-RoPE QK Compression}\label{sec:design:decompress_attn}

Achieving $K$-decompression elimination and attention acceleration requires calculating the common $R$ for post-RoPE $Q_p$ and $K_p$ as indicated in~\cref{sec:motivation}. We explain how to find the common $R$ and use it to compress post-RoPE $Q_p$ and $K_p$.

\begin{figure}[h]\vspace{-0in}
  \centering
  \includegraphics[width=0.99\columnwidth]{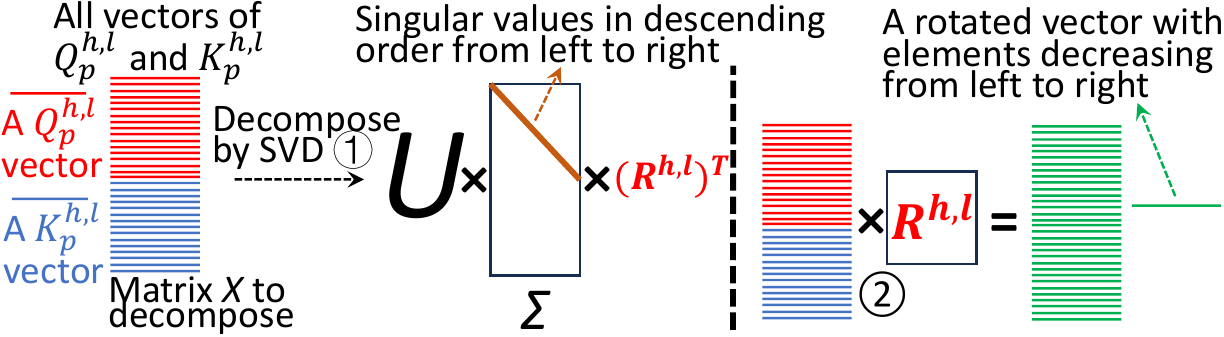}
  \vspace{-0.15in}
  \caption{An example of compressing $Q_p^{h,l}$ and $K_p^{h,l}$.}
  \label{fig:decompose_qk}
  \vspace{-0.15in}
\end{figure}

We begin by generating a series of random tokens and feeding them into the model to extract the post-RoPE representations $Q_p^{h,l}$ and $K_p^{h,l}$ from each head $h$ in every layer $l$. As illustrated in Fig.~\ref{fig:decompose_qk}, for each head $h$ in layer $l$, we concatenate all vectors from $Q_p^{h,l}$ and $K_p^{h,l}$ row-wise to form a matrix $X$. The matrix $X$ is then decomposed using SVD as  $U\Sigma (R^{h,l})^T$ (step \circled{1} in Fig.~\ref{fig:decompose_qk}). The singular values are located on the diagonal of $\Sigma$. The matrix $R^{h,l}$ serves as the rotation matrix used to rotate the vectors in $Q_p^{h,l}$ and $K_p^{h,l}$.
After step \circled{2} in Fig.~\ref{fig:decompose_qk}, the green rotated vectors obtained through $R^{h,l}$ are still arranged row-wise, and each rotated vector has elements arranged in descending order from left to right, a consequence of the singular values' arrangement in $\Sigma$. In other words, the magnitude of the singular values on the diagonal determines the relative size of each element in the rotated vectors. The larger the singular value, the greater the element value in the corresponding dimension, and vice versa. If we want to remove the smallest $N$ elements from each rotated vector, we simply need to delete the rightmost $N$ columns of $R^{h,l}$ before performing step \circled{2} in Fig.~\ref{fig:decompose_qk}, as introduced in~\cref{sec:svd_for_compression}.
Therefore, given a compression rate $p_{QK}^{h,l}$ for $Q_p^{h,l}$ and $K_p^{h,l}$, to compress $Q_p^{h,l}$ and $K_p^{h,l}$, we need to delete the rightmost $p_{QK}^{h,l}d_h$ columns of $R^{h,l}$ before performing step \circled{2}.
\looseness=-1

If the model uses Group Query Attention (GQA), we pair each query head’s $Q_p$ with its key head’s $K_p$ to compute the singular values and the rotation matrix $R$.

\noindent\textbf{How many random tokens do we need?}
The experimental results shown in Fig.~\ref{fig:avg_ratio_diff} help us answer this question. For a given model, we observe that at least 8K random tokens are required to ensure the accuracy of the computed rotation matrix $R$. This is because a larger number of random tokens allows for a more accurate identification of the common low-magnitude dimensions in $Q_p^{h,l}$ and $K_p^{h,l}$.

\noindent\textbf{Does the introduced compression overhead (step \circled{6} in Fig.~\ref{fig:design_overview}) compromise the benefits from decompression elimination and attention acceleration?}
The answer to this question is negative. Experimental results in~\cref{sec:time_overhead} show that the compression overhead is negligible compared to the overall JCT. We analyze why the introduced online compression overhead is small by examining both the prefill and decode phases.
In the prefill phase, the computational complexity of attention is $8sd^2+4s^2d$, while all other linear layers have a complexity of $4sdd_{ff}$~\cite{transformer}, where $d_{ff}$ is the size of the feedforward neural network and larger than $d$. The compression of $Q_p$ and $K_p$ has a complexity of only $4sdd_h$. Since $d_h=128$ is typically 32 to 64 times smaller than $d$, $4sdd_h$ is significantly smaller than $(8sd^2+4s^2d) + (4sdd_{ff}) = 8sd^2+4s^2d + 4sdd_{ff}$. It can be even smaller when the sequence length $s$ is extremely long. This explains why the compression overhead during prefill is negligible.
In the decode phase, the computational complexity of attention is $8d^2+4sd$, while all other linear layers have a complexity of $4dd_{ff}$~\cite{transformer}. The compression of $Q_p$ and $K_p$ has a complexity of only $4dd_h$. $4dd_h$ is also significantly smaller than $(8d^2+4sd) + (4dd_{ff}) = 8d^2+4sd + 4dd_{ff}$, especially when the total sequence length $s$ is extremely long. This explains why the compression overhead during decode is also negligible.

\vspace{-0in}
\subsection{Adaptive Compression Rate Determination}\label{sec:Hybrid}
\vspace{-0in}


Model layers contribute differently to the output~\cite{transformer}, as do the attention heads within each layer~\cite{ge2023model}. This implies that different heads $h$ in each layer $l$ can have different compression rates. How to determine the compression rates for post-RoPE $Q_p^{h,l}$ and $K_p^{h,l}$, as well as $V^{h,l}$ when performing hidden dimension compression remains a question. Since different heads contribute differently, we expect that the distribution of singular values obtained through SVD for post-RoPE $Q_p^{h,l}$-$K_p^{h,l}$ and $V^{h,l}$ may vary across different heads and layers.
The singular values of $V^{h,l}$ are obtained by decomposing the model parameter $W_V^{h,l}$ through SVD~\cite{palu}.
Table~\ref{tab:cv_max_singval_qk} and Table~\ref{tab:cv_max_singval_v} presents the coefficient of variation (CV) of the maximum singular values (i.e., those in the first head dimension) for $Q_p^{h,l}$-$K_p^{h,l}$ and $V^{h,l}$ across all heads of each model with different datasets. The CV is the ratio of the standard deviation to the mean of the maximum singular values of heads. This CV for different models and datasets is around 40\%-50\%, indicating variation in the maximum singular values among different heads. \looseness=-1

\begin{table}[h]\vspace{-0in}
\centering
\small
\begin{tabular}{|c||c|c|c|c|c|}
\hline
& IMDb & arXiv & Cocktail & HumanEval & PG-19 \\
\hline
Mistral-v0.3 7B & 49.76\% & 49.55\% & 49.93\% & 50.04\% & 54.5\% \\
\hline
Llama-3.1 8B & 51.13\% & 51.21\% & 51.07\% & 50.94\% & 52.1\% \\
\hline
Phi-3 14B & 52.97\% & 53.38\% & 52.63\% & 52.53\% & 50.7\% \\
\hline
Yi 34B & 55.21\% & 55.73\% & 54.95\% & 54.98\% & 54.9\% \\
\hline
Llama-3.1 70B & 57.75\% & 58.03\% & 57.69\% & 57.51\% & 59.2\% \\
\hline
\end{tabular}
\vspace{-0in}
\caption{CV of the maximum singular values for $Q_p^{h,l}$-$K_p^{h,l}$.}
\vspace{-0.2in}
\label{tab:cv_max_singval_qk}
\end{table}

\begin{table}[h]\vspace{-0in}
\centering
\small
\begin{tabular}{|c||c|c|c|c|c|}
\hline
& IMDb & arXiv & Cocktail & HumanEval & PG-19 \\
\hline
Mistral-v0.3 7B & 41.86\% & 42.37\% & 41.44\% & 41.31\% & 43.55\%\\
\hline
Llama-3.1 8B & 43.59\% & 43.63\% & 43.56\% & 43.61\% & 42.92\%\\
\hline
Phi-3 14B & 45.37\% & 44.82\% & 44.96\% & 45.69\% & 44.73\%\\
\hline
Yi 34B & 44.48\% & 44.81\% & 44.96\% & 44.52\% & 45.25\%\\
\hline
Llama-3.1 70B & 45.16\% & 45.42\% & 45.38\% & 44.95\% & 45.58\%\\
\hline
\end{tabular}
\vspace{-0in}
\caption{CV of the maximum singular values for $V^{h,l}$.}
\vspace{-0.2in}
\label{tab:cv_max_singval_v}
\end{table}

Since the dominant maximum singular values vary across heads, the compression rate $p$ for each head should also differ. For heads with larger maximum singular values and more dimensions with relatively smaller values, we can use a higher $p$ to remove more low-magnitude dimensions. Conversely, for heads with smaller maximum singular values and fewer dimensions with relatively smaller values, we can use a lower $p$ to remove fewer low-magnitude dimensions. This can achieve a higher overall compression rate $p$ with minimal accuracy loss. How to determine $p$ for $Q_p^{h,l}$-$K_p^{h,l}$ and $V^{h,l}$ in each head? Since singular values represent the magnitude of the components of rotated vectors in their corresponding dimensions, the proportion of the cumulative sum of the removed singular values to the total sum of singular values determines the extent of information loss. We introduce a singular value removal rate $r$ to control this proportion, which is the ratio of the sum of the singular values in the removed dimensions to the total sum of the singular values across all $d_h$ dimensions. In a head, given an $r$, the corresponding $p$ for $Q_p^{h,l}$-$K_p^{h,l}$ and $V^{h,l}$ can be determined. Specifically, given an $r$, we only need to find $p$ that satisfies the following condition:

\vspace{-0in}
\begin{equation}\vspace{-0in}
    \begin{split}
        \frac{\sum_{i=j+1}^{d_h-1} \theta^i}{\sum_{i=0}^{d_h-1} \theta^i} \leq r < \frac{\sum_{i=j}^{d_h-1} \theta^i}{\sum_{i=0}^{d_h-1} \theta^i} \mbox{ AND } j = \lfloor (1-p)(d_h-1)\rfloor,
    \end{split}
\end{equation}

\noindent where $\theta^i$ is the singular value at head dimension $i$. We set a shared $r$ for all heads across different layers, allowing us to directly determine the $p$ required for $Q_p^{h,l}$-$K_p^{h,l}$ and $V^{h,l}$ in each head, thus obtaining an overall KV compression rate, which corresponds to an accuracy constraint. This enables us to construct a function $F_p$ that takes an accuracy constraint as input and outputs a combination of $p$ needed for all heads under this constraint. During inference, given an accuracy constraint, we use $F_p$ to directly obtain the $p$ needed for $Q_p^{h,l}$-$K_p^{h,l}$ and $V^{h,l}$ in each head.

\subsection{Kernel Enhancement for Adaptive Compression Rate}\label{sec:kernel_opt}

\begin{wrapfigure}[7]{c}{0.27\textwidth}\vspace{-0.18in}
  \centering
  \includegraphics[width=0.27\textwidth]{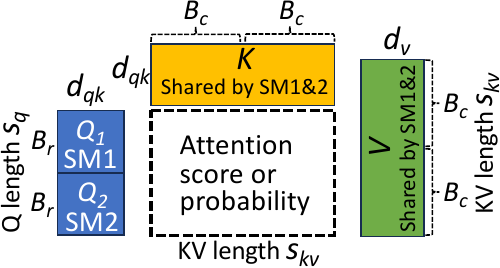}
  \vspace{-0.37in}
  \caption{Attention kernel.}
  \label{fig:attn_kernel}
\end{wrapfigure}

State-of-the-art memory-efficient attention kernel implementations (e.g., FlashAttention-2~\cite{flashattn2}) operate as shown in Fig.~\ref{fig:attn_kernel}.
The attention computation of a batch of requests is divided into $batch\_size\times N_h$ computation units, each corresponding to a head of a request.
Each unit further splits $Q$ into blocks of shape ($B_r=64$, $d_{qk}=d_h$), where $B_r$ is the row block size and $d_{qk}$ is the head dimension size for $Q$ and $K$. Each $Q$ block is processed on one SM as an SM task. All SMs responsible for a head share that head's $K$ and $V$ to minimize data movement between SMs, thereby reducing memory access overhead. When an SM processes a $Q$ block, $K$, and $V$, it further divides $K$ and $V$ into blocks of shape ($B_c=64$, $d_{v}=d_h$) to be processed sequentially in time order, where $B_c$ is the column block size and $d_v$ is the head dimension size for $V$. Without adaptive compression rates, all heads use the same $d_{qk}$ and the same $d_v$. However, with adaptive compression rates, different heads have different $d_{qk}$ and different $d_v$, resulting in varying computational loads and uneven execution times for each SM. The kernel runtime is determined by the SM with the longest execution time. Therefore, it is necessary to enhance the attention kernel with adaptive compression rates. We propose different enhancements for prefill and decode. \looseness=-1


\setlength{\columnsep}{8pt}
\noindent\textbf{Prefill.}
During prefill, the computational workload of an SM task includes $2B_rd_{qk}s_{kv}$ for multiplying the $Q$ block with $K$, $2B_rs_{kv}$ for softmax, and $2B_rs_{kv}d_v$ for multiplying attention probability with $V$, resulting in a total of $2B_rs_{kv}(d_{qk}+d_v+1)$, where $s_{kv}$ is the sequence length of $K$ and $V$. Let $d_{sum}^i=d_{qk}+d_v+1$ for head $i$ and $d_{sum}^{max}$ be the maximum $d_{sum}^i$ across all heads. As the compression rate changes, $d_{sum}^i$ changes accordingly. We increase $B_r$ to $\frac{d_{sum}^{max}}{d_{sum}^i}B_r$ for head $i$, ensuring that the computational workload for each SM task is the same, preventing uneven execution times among SMs.
By increasing $B_r$, the number of $Q$ blocks per head decreases, reducing the total number of SM tasks. As a result, the GPU can complete all SM tasks in fewer rounds, with each round using all SMs of the GPU.
In the final round, not all SMs may be fully utilized. To maximize the utilization, we redistribute the remaining SM tasks in the last round. Suppose the GPU has $N_{SM}$ SMs and the last round has $N_{last}$ SM tasks. The workload assigned to each SM should be $wl=\frac{1}{N_{SM}}\sum_{i=1}^{N_{last}}B_r^{tsk(i)}d_{sum}^{tsk(i)}$, where $B_r^{tsk(i)}$ and $d_{sum}^{tsk(i)}$ is $B_r$ and $d_{sum}$ of the $i$-th SM task in the last round. Then, task $i$ should be split into smaller blocks using new row block size $wl/d_{sum}^{tsk(i)}$. All redistributed tasks in the last round are sequentially assigned to an SM with the smallest allocated workloads one at a time.
\setlength{\columnsep}{\defaultcolumnsep}

\begin{wrapfigure}[5]{c}{0.25\textwidth}\vspace{-0.2in}
  \centering
  \includegraphics[width=0.25\textwidth]{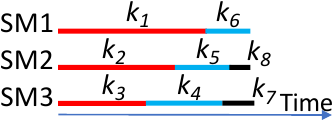}
  \vspace{-0.35in}
  \caption{Task allocation.}
  \label{fig:task_allocation}
\end{wrapfigure}
\setlength{\columnsep}{8pt}
\noindent\textbf{Decode.}
In decode, the Q length $s_q$ for each head is only 1, meaning it cannot be divided into multiple $Q$ blocks. Consequently, each head corresponds to one SM task, and adjusting $B_r$ to enhance the kernel is not feasible. Instead, we schedule these tasks in a heuristic way to avoid heavy overhead. The execution time of each task $k_i$ is determined by $s_{kv}\times d_{sum}$. All tasks are sorted in descending order of execution time and then sequentially assigned to an SM with the smallest cumulative assigned time one at a time. Fig.~\ref{fig:task_allocation} provides an example where 7 tasks are distributed across 3 SMs. Tasks $k_1$-$k_7$ are sorted in descending order of execution time. Following our scheduling strategy, $k_1$ and $k_6$ are assigned to SM1, $k_2$, $k_5$, and $k_8$ to SM2, and $k_3$, $k_4$, and $k_7$ to SM3.
\setlength{\columnsep}{\defaultcolumnsep}

\vspace{-0.1in}
\section{Performance Evaluation}\label{sec:exp}
\vspace{-0in}
\subsection{Implementation}
\vspace{-0in}

We implemented \sys based on vLLM~\cite{vllm2023kwon} and modified vLLM model classes to enable QKV compression and adaptive compression rates. We used NVIDIA CUTLASS~\cite{cutlass} to implement the on-device matrix multiplication for the attention kernel. A load balancer was implemented with a fast Python web framework Quart~\cite{quart}, which allocates each request to a vLLM instance with the shortest queue length.

\noindent\textbf{Kernel enhancement for prefill.} For prefill-phase kernel enhancement (\cref{sec:kernel_opt}), a large $B_r$ can cause data to exceed the register capacity of a single SM. In such cases, we reduce $B_c$ to prevent this issue, which does not affect the total computational workload of an SM task since $s_{kv}$ remains unchanged. Limited by the tensor core interface used for matrix operations, the values of $d_{qk}$, $d_v$, $B_r$, and $B_c$ must be multiples of 16. Therefore, these values are rounded up to the nearest multiple of 16, and any padding required is filled with zeros. For a given overall KV compression rate $p$, the $d_{sum}$ and $B_c$ for each head are determined, and $B_r$ used by each head is also determined except in the final round. In the last round, $B_r$ can be enumerated among multiples of 16. For different overall KV compression rates, we pre-compile all kernels offline according to the required matrix shapes.

\noindent\textbf{Kernel enhancement for decode.} For decode-phase kernel enhancement, we fused the scheduling computations (a significant portion of which involves sorting) to a CUDA kernel, thereby minimizing the time overhead.

\subsection{Setup}


In the evaluation, unless otherwise specified, we utilized the following settings. Table~\ref{tab:models} lists the state-of-the-art models we used with their Tensor Parallelism (TP) and Pipeline Parallelism (PP) size, following the setting in~\cite{Agrawal2023SARATHIEL, distserve}.
We used the datasets listed in Table~\ref{tab:datasets}. IMDb includes 27 genres of movies, TV shows, etc., collected on the Internet. It is operated by IMDb.com, Inc., a subsidiary of Amazon. The arXiv summarization has a collection of scientific publications and their summaries on arXiv.org~\cite{arxiv}. Information Retrieval (IR) is the process of retrieving relevant content from vast amounts of information based on a user query. The Cocktail is a benchmark for IR, including 8 different IR tasks such as question answering, fact checking, etc. HumanEval evaluates the performance of code completion, including 164 programming problems.
The Google PG-19 dataset provides a description of a book passage and requires the model to generate a book of a specified length.
For the summarization and book generation tasks, we use \textit{ROUGE-1}~\cite{rouge-score} as the accuracy metric; for the code completion task, we use \textit{Edit Similarity (normalized Levenshtein distance)}~\cite{zhang2024hierarchicalcontextpruningoptimizing, string-similarity}.

\begin{table}[h]\vspace{-0in}
\centering
\small
\begin{adjustbox}{max width=\columnwidth}
\begin{tabular}{|c|c||c|c|}
\hline
Mistral-v0.3 7B~\cite{mistral-v0.3} & no TP and PP & Llama-3.1 8B~\cite{llama3.1} & no TP and PP \\
\hline
Phi-3 14B~\cite{phi-3} & no TP and PP & Yi 34B~\cite{yi-model} & TP=4 \\
\hline
Llama-3.1 70B~\cite{llama3.1} & TP=4 & & \\
\hline
\end{tabular}
\end{adjustbox}
\vspace{-0in}
\caption{Model size and TP/PP size.}
\vspace{-0.23in}
\label{tab:models}
\end{table}

\begin{table}[h]\vspace{-0.1in}
\centering
\small
\begin{tabular}{l|lll|lll|}
\hline
\multicolumn{1}{|l|}{Trace} &
  \multicolumn{3}{l|}{Input length} &
  \multicolumn{3}{l|}{Output length} \\
  \cline{1-7}
  \multicolumn{1}{|l|}{} &
  \multicolumn{1}{l|}{avg} &
  \multicolumn{1}{l|}{min} &
  max &
  \multicolumn{1}{l|}{avg} &
  \multicolumn{1}{l|}{min} &
  max \\ \hline
\multicolumn{1}{|l|}{IMDb classification~\cite{imdb}}     & \multicolumn{1}{l|}{315}  & \multicolumn{1}{l|}{106}  & 821 & \multicolumn{1}{l|}{37}  & \multicolumn{1}{l|}{16} & 87  \\ \hline
\multicolumn{1}{|l|}{arXiv summarization~\cite{arxiv-summarization}}   & \multicolumn{1}{l|}{6.3K}  & \multicolumn{1}{l|}{1.6K} & 14.1K  & \multicolumn{1}{l|}{243} & \multicolumn{1}{l|}{29} & 464   \\ \hline
\multicolumn{1}{|l|}{Cocktail for IR~\cite{cocktailforir}}   & \multicolumn{1}{l|}{49.7K}  & \multicolumn{1}{l|}{9.4K} & 104.9K  & \multicolumn{1}{l|}{159} & \multicolumn{1}{l|}{44} & 246   \\ \hline
\multicolumn{1}{|l|}{HumanEval~\cite{humaneval}}   & \multicolumn{1}{l|}{204}  & \multicolumn{1}{l|}{75} & 697  & \multicolumn{1}{l|}{139} & \multicolumn{1}{l|}{11} & 552   \\ \hline
\multicolumn{1}{|l|}{Google PG-19~\cite{pg-19}}   & \multicolumn{1}{l|}{277}  & \multicolumn{1}{l|}{138} & 419  & \multicolumn{1}{l|}{61.5K} & \multicolumn{1}{l|}{24.3K} & 97.5K   \\ \hline
\end{tabular}%
\vspace{-0in}
\caption{Trace properties.}
\label{tab:datasets}
\vspace{-0.23in}
\end{table}

\DEL{\begin{figure}
  \centering
  \includegraphics[width=0.22\textwidth]{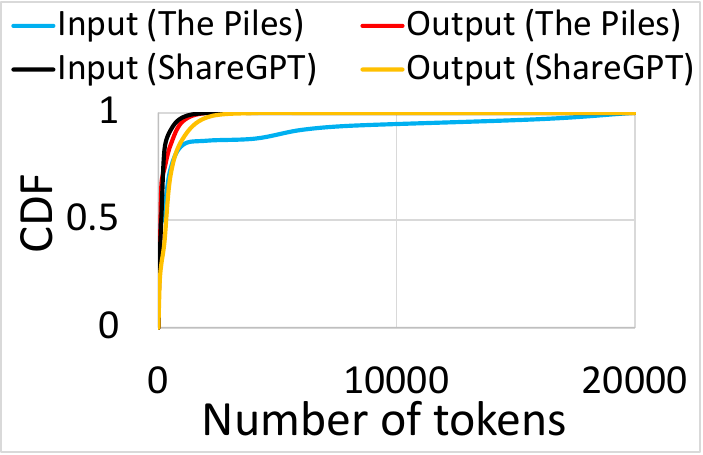}
  \caption{CDF of examples v.s. the number of tokens. \sh{add X values in the curve point}}
  \label{fig:cdf_sequences}
\end{figure}}



We employed four AWS p4de.24xlarge instances \cite{awsp4} located in four nodes. Each instance is equipped with 8 NVIDIA A100 GPUs (each with 80 GiB memory), 96 vCPUs, and 1152 GiB host memory, connected with a 400 Gbps network. As in \cite{holmes2024deepspeedfastgen}, we executed 16 concurrent clients to dispatch requests from a single dataset. We built our system atop vLLM~\cite{vllm2023kwon} and scheduled each request to a vLLM instance (which holds one model replica) that has the shortest queue length~\cite{splitwise}, defined by the number of tokens. By default, the request rate is set to the maximum processing capacity without increasing queuing time, based on a Poisson distribution. The batch size is automatically determined by vLLM based on the available space in the KV cache.\looseness=-1

\vspace{-0in}
\subsection{Compared Methods}
\vspace{-0.05in}


We selected Palu as our comparison method. \sys and Palu compress the hidden dimension of KV data, which complements token eviction-based and quantization-based methods. Thus, both \sys and Palu can be combined with them to improve compression rates further~\cite{palu}.
We selected the most representative state-of-the-art eviction-based method, Keyformer~\cite{keyformer}, and quantization-based method, KVQuant~\cite{kvquant}.
We combined \sys with Keyformer and KVQuant to create enhanced versions, referred to as KF-F and KVQ-F, respectively. We also combined Palu with Keyformer and KVQuant, denoted by KF-P and KVQ-P, respectively.

\noindent\textbf{KF-F and FK-P.}
KF-F and KF-P apply Keyformer to evict KV and then use \sys and Palu to obtain compressed KV, respectively. Subsequent steps remain the same as \sys and Palu. 

\noindent\textbf{KVQ-F and KVQ-P.}
KVQ-F and KVQ-P first use \sys and Palu to obtain compressed KV, respectively, and then apply KVQuant to quantize the compressed KV. Before \sys's attention and Palu decompressing $K$, the quantized KV is dequantized. Subsequent steps remain the same as \sys and Palu.


\begin{table}[h]\vspace{-0in}
\centering
\begin{tabular}{|c||c|c||c|c||c|c|}
\hline
& \sys & Palu & KF-F & KF-P & KVQ-F & KVQ-P \\
\hline
Mistral-v0.3 7B &  0.61 & 0.55 & 0.71 & 0.68 & 0.90 & 0.88 \\
\hline
Llama-3.1 8B &  0.53 & 0.51 & 0.72 & 0.69 & 0.89 & 0.88 \\
\hline
Phi-3 14B &  0.58 & 0.50 & 0.80 & 0.75 & 0.92 & 0.90 \\
\hline
Yi 34B &  0.55 & 0.49 & 0.75 & 0.71 & 0.92 & 0.91 \\
\hline
Llama-3.1 70B &  0.63 & 0.57 & 0.74 & 0.72 & 0.91 & 0.89 \\
\hline
\end{tabular}
\vspace{-0in}
\caption{Overall KV compression rate (IMDb).}
\vspace{-0.23in}
\label{tab:kv_compress_rate_imdb}
\end{table}

\begin{table}[h]\vspace{-0in}
\centering
\begin{tabular}{|c||c|c||c|c||c|c|}
\hline
& \sys & Palu & KF-F & KF-P & KVQ-F & KVQ-P \\
\hline
Mistral-v0.3 7B &  0.64 & 0.59 & 0.84 & 0.81 & 0.92 & 0.90 \\
\hline
Llama-3.1 8B &  0.65 & 0.55 & 0.84 & 0.82 & 0.90 & 0.88 \\
\hline
Phi-3 14B &  0.62 & 0.55 & 0.82 & 0.80 & 0.91 & 0.89 \\
\hline
Yi 34B &  0.64 & 0.57 & 0.87 & 0.84 & 0.91 & 0.90 \\
\hline
Llama-3.1 70B &  0.67 & 0.58 & 0.86 & 0.82 & 0.92 & 0.90 \\
\hline
\end{tabular}
\vspace{-0in}
\caption{Overall KV compression rate (arXiv).}
\vspace{-0.23in}
\label{tab:kv_compress_rate_arxiv}
\end{table}

\begin{table}[h]\vspace{-0in}
\centering
\begin{tabular}{|c||c|c||c|c||c|c|}
\hline
& \sys & Palu & KF-F & KF-P & KVQ-F & KVQ-P \\
\hline
Mistral-v0.3 7B &  0.58 & 0.51 & 0.79 & 0.75 & 0.90 & 0.89 \\
\hline
Llama-3.1 8B &  0.56 & 0.50 & 0.75 & 0.70 & 0.91 & 0.89 \\
\hline
Phi-3 14B &  0.67 & 0.59 & 0.82 & 0.77 & 0.92 & 0.90 \\
\hline
Yi 34B &  0.60 & 0.55 & 0.83 & 0.78 & 0.92 & 0.91 \\
\hline
Llama-3.1 70B &  0.59 & 0.52 & 0.84 & 0.76 & 0.92 & 0.89 \\
\hline
\end{tabular}
\vspace{-0in}
\caption{Overall KV compression rate (Cocktail).}
\vspace{-0.23in}
\label{tab:kv_compress_rate_cocktail}
\end{table}

\begin{table}[h]\vspace{-0in}
\centering
\begin{tabular}{|c||c|c||c|c||c|c|}
\hline
& \sys & Palu & KF-F & KF-P & KVQ-F & KVQ-P \\
\hline
Mistral-v0.3 7B &  0.53 & 0.50 & 0.73 & 0.69 & 0.89 & 0.89 \\
\hline
Llama-3.1 8B &  0.54 & 0.48 & 0.76 & 0.72 & 0.87 & 0.89 \\
\hline
Phi-3 14B &  0.54 & 0.50 & 0.77 & 0.72 & 0.91 & 0.90 \\
\hline
Yi 34B &  0.59 & 0.53 & 0.78 & 0.73 & 0.90 & 0.91 \\
\hline
Llama-3.1 70B &  0.56 & 0.52 & 0.77 & 0.74 & 0.87 & 0.89 \\
\hline
\end{tabular}
\vspace{-0in}
\caption{Overall KV compression rate (HumanEval).}
\vspace{-0.23in}
\label{tab:kv_compress_rate_humaneval}
\end{table}

\begin{table}[h]\vspace{-0in}
\centering
\begin{tabular}{|c||c|c||c|c||c|c|}
\hline
& \sys & Palu & KF-F & KF-P & KVQ-F & KVQ-P \\
\hline
Mistral-v0.3 7B &  0.65 & 0.58 & 0.81 & 0.77 & 0.92 & 0.91 \\
\hline
Llama-3.1 8B &  0.63 & 0.56 & 0.83 & 0.80 & 0.91 & 0.89 \\
\hline
Phi-3 14B &  0.66 & 0.59 & 0.79 & 0.74 & 0.91 & 0.90 \\
\hline
Yi 34B &  0.65 & 0.57 & 0.78 & 0.80 & 0.92 & 0.91 \\
\hline
Llama-3.1 70B &  0.67 & 0.59 & 0.83 & 0.79 & 0.91 & 0.89 \\
\hline
\end{tabular}
\vspace{-0in}
\caption{Overall KV compression rate (PG-19).}
\vspace{-0.23in}
\label{tab:kv_compress_rate_pg19}
\end{table}

\vspace{-0.15in}
\subsection{Comparison of Compression Rates}
\vspace{-0.05in}


We study the overall KV compression rate $p$ given an accuracy constraint for different methods. Specifically, we focus on the scenarios where the methods achieve 99\% of the baseline accuracy that is achieved without compression~\cite{keyformer}.
For KF-F, we configure the \sys and Keyformer components to contribute equally to the overall compression rate.
KF-P follows the same way.
Since KVQuant can only be configured with 2-4 quantization bit widths, we use 4-bit quantization for KVQuant, KVQ-F, and KVQ-P to ensure the accuracy can reach 99\% of the baseline accuracy without compression. We use binary search to determine the maximum $p$ that achieves 99\% of the baseline accuracy. Table~\ref{tab:kv_compress_rate_imdb}-\ref{tab:kv_compress_rate_pg19} show the results for different datasets. \sys alone achieves $p$ of 0.53-0.67 across all datasets, while Palu achieves around 0.06 smaller $p$ than \sys.
KF-F and KVQ-F have around 0.04 and 0.01 higher $p$ than KF-P and KVQ-P, respectively. The improvement stems from the adaptive compression rate introduced in \sys.
KF-F, KVQ-F, KV-P, and KVQ-P all have higher compression rates than \sys and Palu, indicating the enhanced version of Keyformer and KVQuant can improve the compression rates further.

\vspace{-0.1in}
\subsection{Overall System Performance}
\vspace{-0.05in}

We tested Mistral-v0.3 7B, Phi-3 14B, Yi 34B, and Llama-3.1 70B, denoted by M, P, Y, and L. Since IMDb and HumanEval are both datasets with short sequences, we kept the popular dataset, HumanEval, and excluded IMDb.
The KV compression rates for them follow those in Table~\ref{tab:kv_compress_rate_arxiv}-\ref{tab:kv_compress_rate_pg19} under the same accuracy constraint. \looseness=-1

\begin{figure}[h]\vspace{-0.1in}
    \centering
    \subfigure[arXiv summarization.\label{fig:avg_ttft_arxiv}]
    {\includegraphics[width=0.495\columnwidth,height=2.3cm]{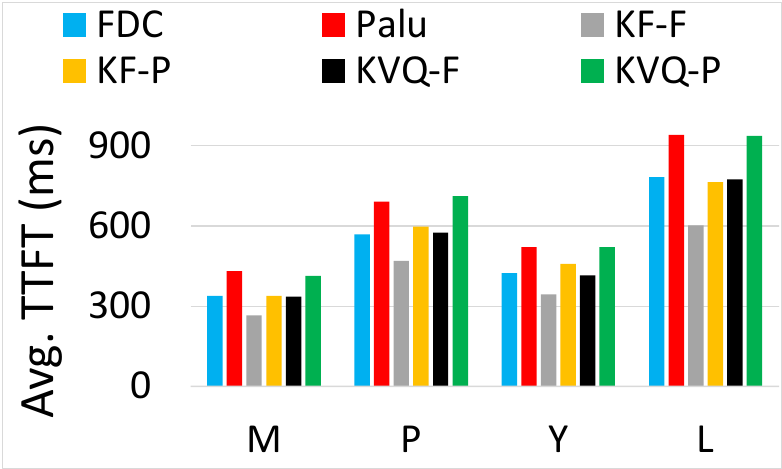}}
    \subfigure[Cocktail dataset.\label{fig:avg_ttft_cocktail}]
    {\includegraphics[width=0.495\columnwidth,height=2.3cm]{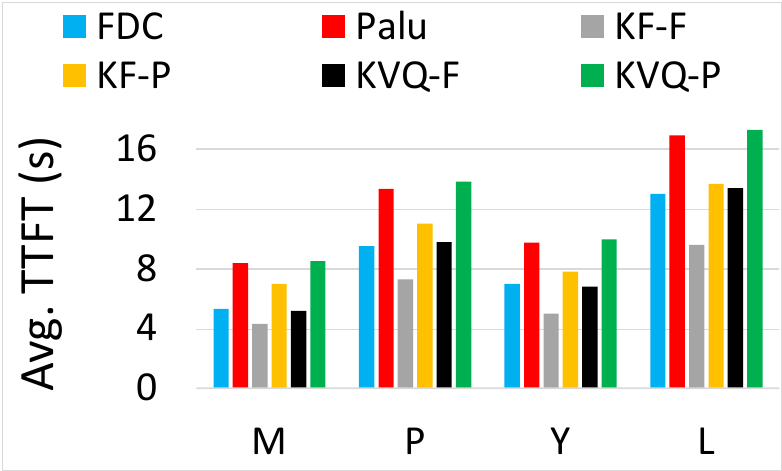}}

    \vspace{-0.1in}
    \subfigure[HumanEval.\label{fig:avg_ttft_humaneval}]
    {\includegraphics[width=0.495\columnwidth,height=2.3cm]{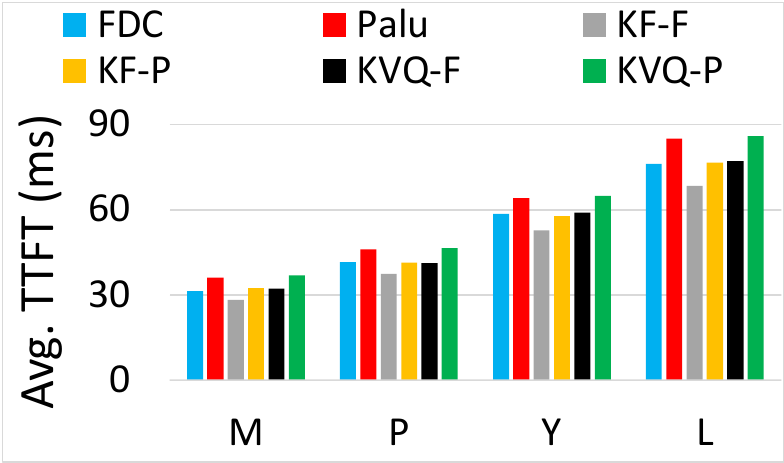}}
    \subfigure[PG-19 dataset.\label{fig:avg_ttft_pg19}]
    {\includegraphics[width=0.495\columnwidth,height=2.3cm]{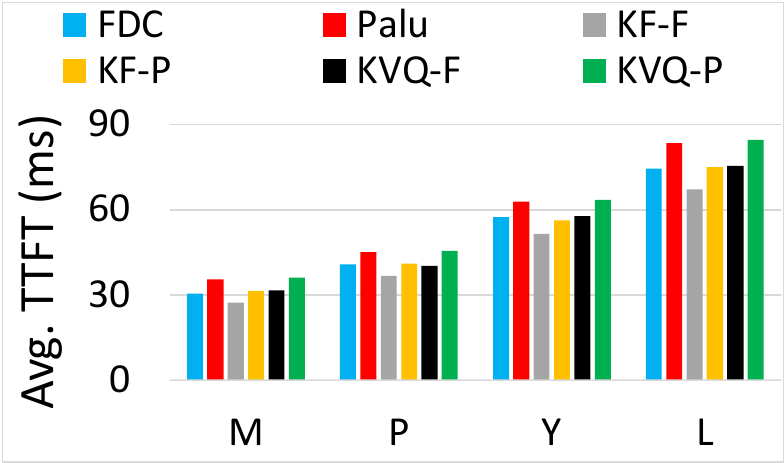}}
    \vspace{-0.25in}
    \caption{Average TTFT under different datasets.}
    \label{fig:avg_ttft} \vspace{-0.25in}
\end{figure}

\begin{figure}[h]\vspace{-0in}
    \centering
    \subfigure[arXiv summarization.\label{fig:avg_tbt_arxiv}]
    {\includegraphics[width=0.495\columnwidth,height=2.3cm]{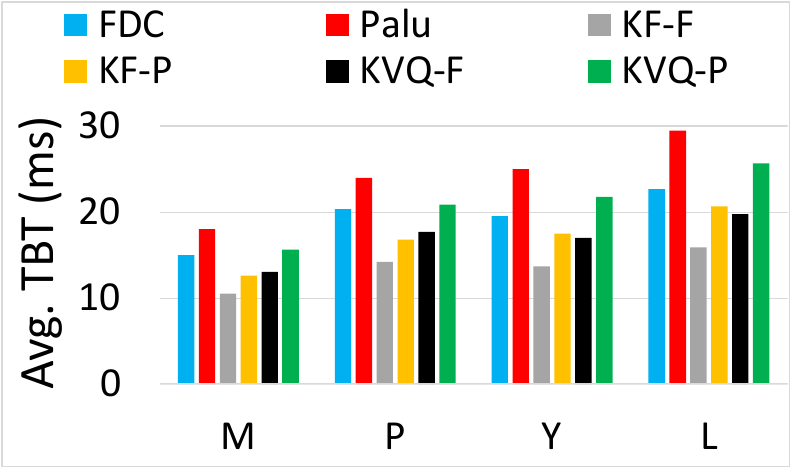}}
    \subfigure[Cocktail dataset.\label{fig:avg_tbt_cocktail}]
    {\includegraphics[width=0.495\columnwidth,height=2.3cm]{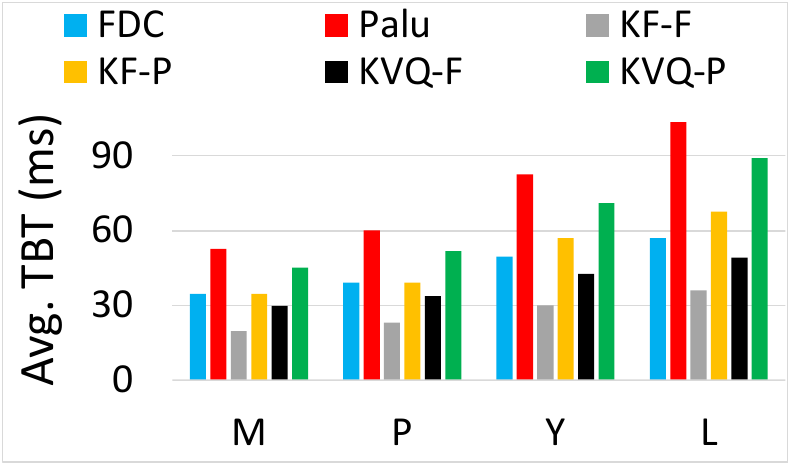}}

    \vspace{-0.1in}
    \subfigure[HumanEval.\label{fig:avg_tbt_humaneval}]
    {\includegraphics[width=0.495\columnwidth,height=2.3cm]{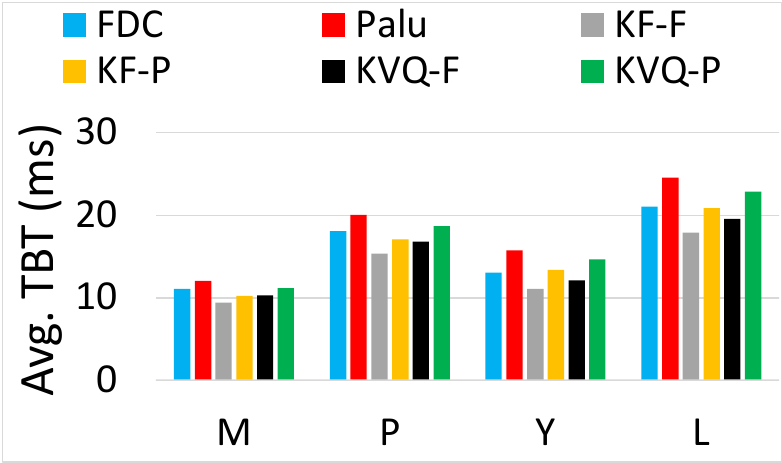}}
    \subfigure[PG-19 dataset.\label{fig:avg_tbt_pg19}]
    {\includegraphics[width=0.495\columnwidth,height=2.3cm]{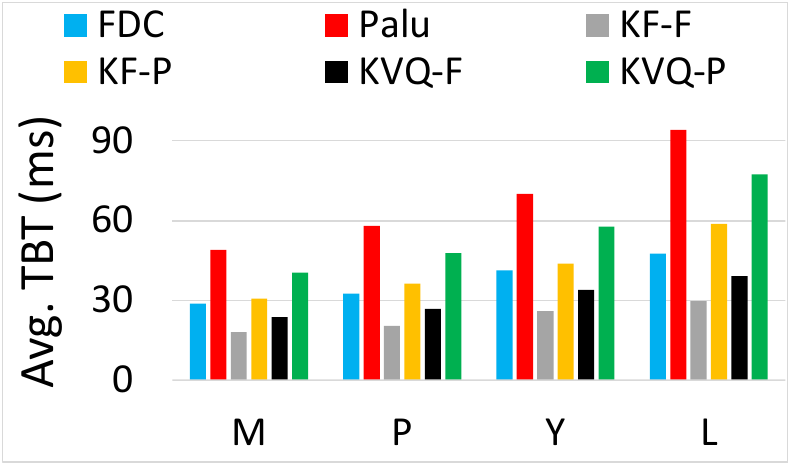}}
    \vspace{-0.25in}
    \caption{Average TBT under different datasets.}
    \label{fig:avg_tbt} \vspace{-0.1in}
\end{figure}

\noindent\textbf{TTFT.}
Fig.~\ref{fig:avg_ttft} shows the average TTFT over requests for different models and datasets. For arXiv, Cocktail, HumanEval, and PG-19, \sys reduces TTFT by 17\%-21\%, 20\%-35\%, 8\%-13\%, and 9\%-12\%, respectively, compared to Palu. This is because \sys not only reduces the $K$-decompression overhead but also computes on compressed data during attention, accelerating the attention computation. Additionally, \sys leverages adaptive compression rates and attention kernel enhancement to further reduce the attention time. The improvement is particularly significant for arXiv and Cocktail, as these datasets have longer input lengths, meaning that attention time constitutes a higher proportion of TTFT. Consequently, the reduction in attention time contributes more to the overall TTFT reduction. We observe a similar trend between KF-F and KF-P and between KVQ-F and KVQ-P, respectively. These improvements arise from the benefits provided by \sys, as described before. For a larger model, the improvement of \sys tends to be smaller because the proportion of attention time in TTFT tends to decrease.

\noindent\textbf{TBT.}
Fig.~\ref{fig:avg_tbt} shows the average TBT over requests for different models and datasets. For arXiv, Cocktail, HumanEval, and PG-19, \sys reduces TBT by 16\%-24\%, 29\%-54\%, 11\%-17\%, and 46\%-64\%, respectively, compared to Palu. In addition to decompression overhead and attention acceleration, the improvement in TBT also stems from the reduced memory overhead due to the smaller KV size.
There is a similar trend between KF-F and KF-P and between KVQ-F and KVQ-P, respectively. For a larger model, the improvement of \sys regarding TBT tends to increase because the proportion of $K$-decompression overhead in decode time tends to increase, which is eliminated by \sys.
As shown in Fig.~\ref{fig:diff_output_length} and Fig.~\ref{fig:diff_model_long_output}, the JCT for sequences with long output lengths is predominantly determined by the decode time. Therefore, for the long-output dataset PG-19, \sys achieves up to a 64\% reduction in JCT compared to Palu.

\begin{figure*}[h]
  \centering
  \includegraphics[width=1\textwidth]{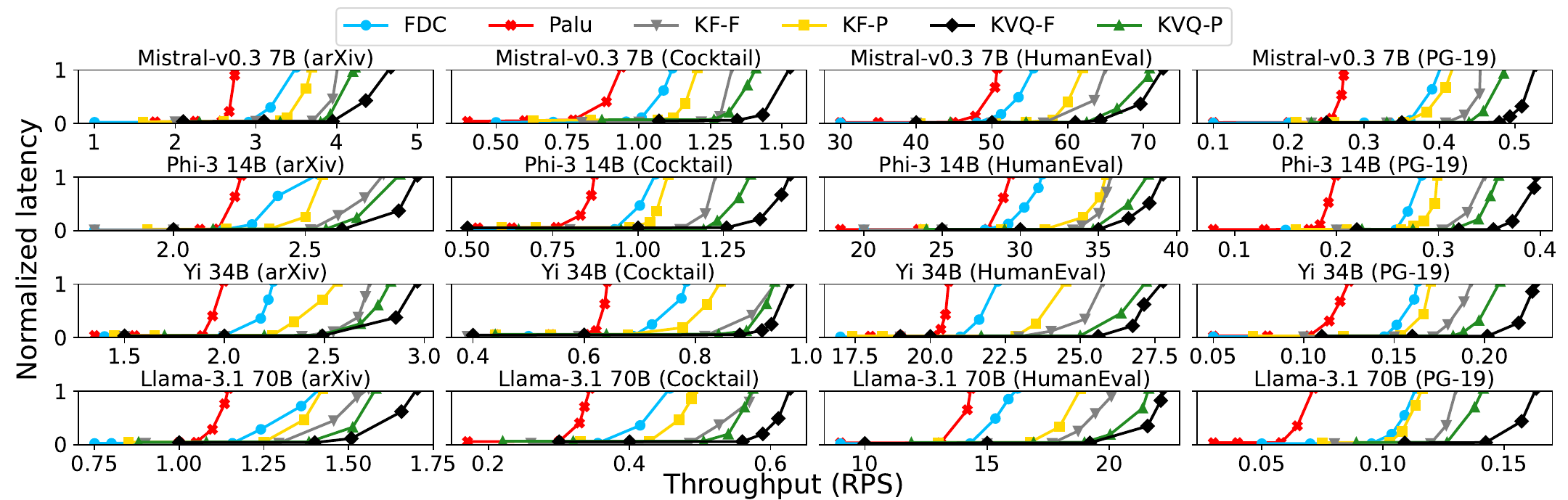}
  \vspace{-0.35in}
  \caption{Normalized latency under different throughput (RPS).}
  \vspace{-0.1in}
  \label{fig:norm_latency}
\end{figure*}


\noindent\textbf{Throughput study.}
We vary the overall throughput, Request Per Second (RPS), and record the normalized latency for each output token~\cite{vllm2023kwon}, calculated by dividing the end-to-end time (i.e., JCT) by the number of output tokens. Fig.~\ref{fig:norm_latency} shows the normalized latency versus RPS for all models and datasets. When the RPS exceeds capacity, the queue length grows faster than the processing speed, leading to an infinite increase in queuing time and latency.
Compared to Palu, \sys achieves 1.18-1.35$\times$, 1.42-1.67$\times$, 1.09-1.21$\times$, and 1.53-1.97$\times$ RPS for arXiv, Cocktail, HumanEval, and PG-19, respectively, while preserving the same latency. The improvement is high for long-sequence datasets Cocktail and PG-19, because \sys eliminates the amplified decompression overhead and reduces the attention time. Additionally, the higher the KV compression rate, the higher the maximum RPS that can maintain a constant low latency. \sys benefits from the adaptive compression rates to have a higher overall compression rate and throughput.
We observe a similar trend between KF-F and KF-P and between KVQ-F and KVQ-P, respectively, due to the same reason.


\vspace{-0in}
\subsection{Ablation Study}

We test the variants of \sys as follows to evaluate each individual method.
1) \sys/PRC is \sys without Post-RoPE QK Compression for decompression overhead elimination and attention acceleration. It compresses $K$ before RoPE and uses uncompressed $K$ in attention, following Palu.
2) \sys/AC is \sys without Adaptive Compression rate. It uses the same compression rate for all heads to achieve 99\% of the accuracy without compression.
3) \sys/KE is \sys without the attention Kernel Enhancement.

\begin{figure}[h]\vspace{-0in}
    \centering
    \subfigure[arXiv summarization.\label{fig:avg_ttft_indiv_arxiv}]
    {\includegraphics[width=0.495\columnwidth,height=2.3cm]{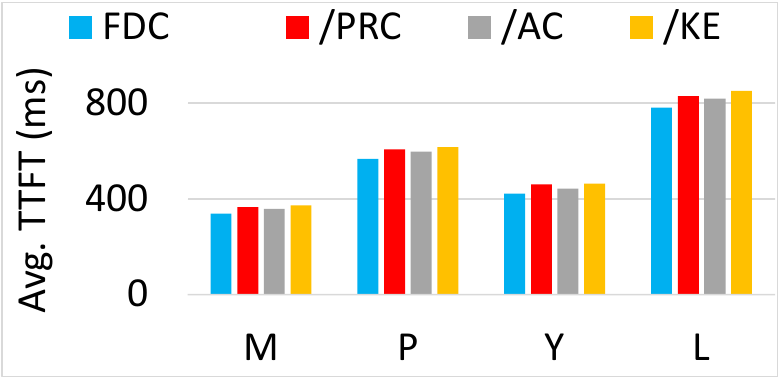}}
    \subfigure[Cocktail dataset.\label{fig:avg_ttft_indiv_cocktail}]
    {\includegraphics[width=0.495\columnwidth,height=2.3cm]{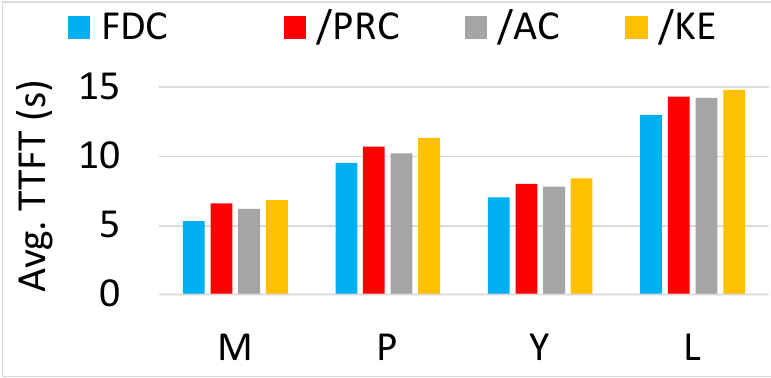}}

    \vspace{-0.1in}
    \subfigure[HumanEval.\label{fig:avg_ttft_indiv_humaneval}]
    {\includegraphics[width=0.495\columnwidth,height=2.3cm]{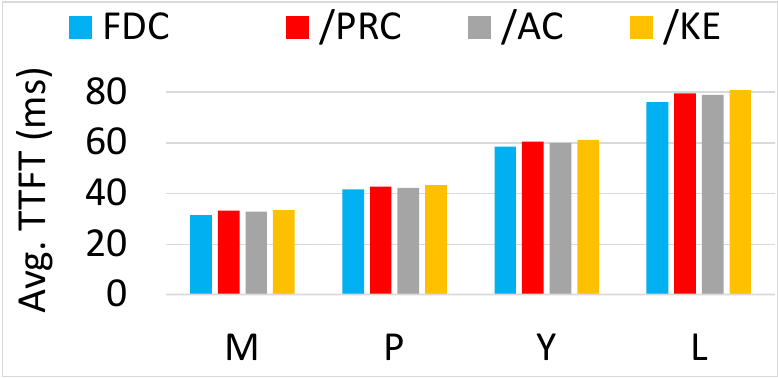}}
    \subfigure[PG-19 dataset.\label{fig:avg_ttft_indiv_pg19}]
    {\includegraphics[width=0.495\columnwidth,height=2.3cm]{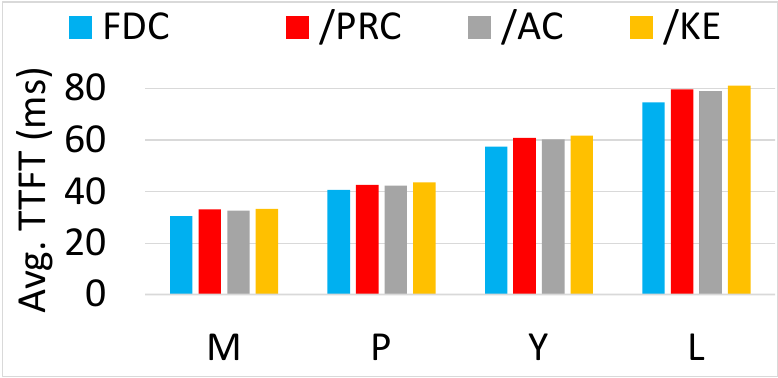}}
    \vspace{-0.25in}
    \caption{Average TTFT for individual methods.}
    \label{fig:avg_ttft_indiv} \vspace{-0.2in}
\end{figure}

\begin{figure}[h]\vspace{-0in}
    \centering
    \subfigure[arXiv summarization.\label{fig:avg_tbt_indiv_arxiv}]
    {\includegraphics[width=0.495\columnwidth,height=2.3cm]{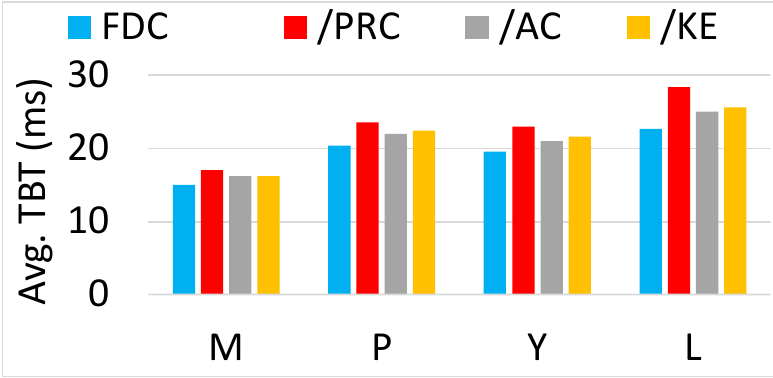}}
    \subfigure[Cocktail dataset.\label{fig:avg_tbt_indiv_cocktail}]
    {\includegraphics[width=0.495\columnwidth,height=2.3cm]{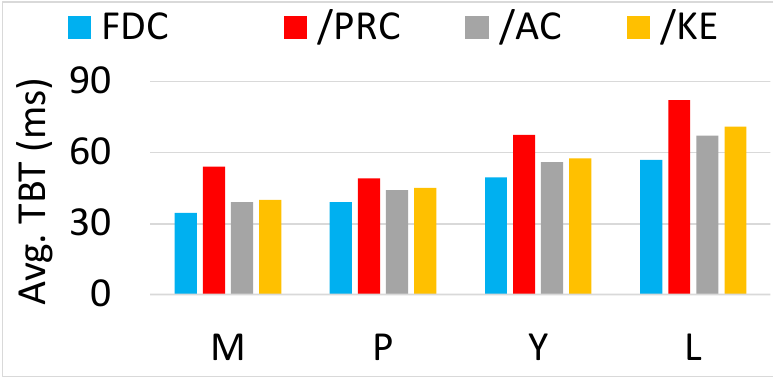}}

    \vspace{-0.1in}
    \subfigure[HumanEval.\label{fig:avg_tbt_indiv_humaneval}]
    {\includegraphics[width=0.495\columnwidth,height=2.3cm]{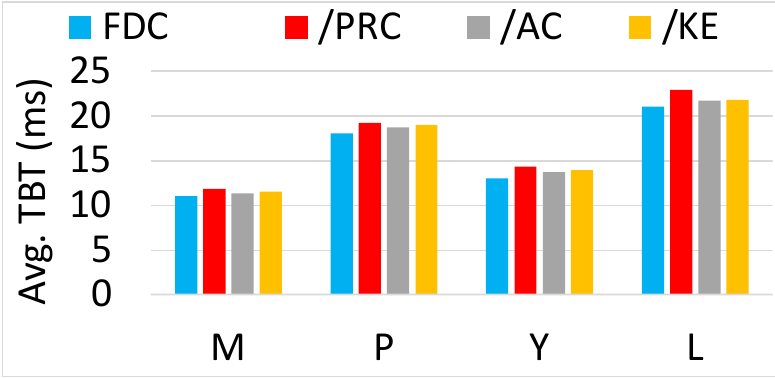}}
    \subfigure[PG-19 dataset.\label{fig:avg_tbt_indiv_pg19}]
    {\includegraphics[width=0.495\columnwidth,height=2.3cm]{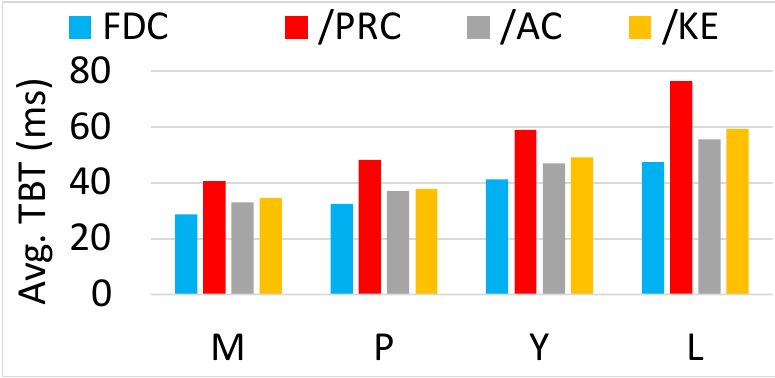}}
    \vspace{-0.25in}
    \caption{Average TBT for individual methods.}
    \label{fig:avg_tbt_indiv} \vspace{-0.2in}
\end{figure}

Fig.~\ref{fig:avg_ttft_indiv} and Fig.~\ref{fig:avg_tbt_indiv} show the average TTFT and TBT over requests for individual methods with different models and datasets. As shown in Fig.~\ref{fig:avg_ttft_indiv}, /PRC has 5\%-9\%, 6\%-17\%, 4\%-5\%, and 4\%-6\% higher TTFT than \sys for arXiv, Cocktail, HumanEval, and PG-19 because /PRC performs $K$-decompression online and uses uncompressed data for attention computation.
/AC has 3\%-8\%, 4\%-10\%, 3\%-5\%, and 3\%-5\% higher TTFT than \sys because, without the adaptive compression rate, the overall KV compression rate required to achieve 99\% of the accuracy without compression is lower, thereby increasing attention time.
/KE has 3\%-9\%, 4\%-11\%, 3\%-6\%, and 4\%-5\% higher TTFT than \sys due to the uneven computation times of SMs.
In Fig.~\ref{fig:avg_tbt_indiv}, /PRC has 11\%-18\%, 29\%-34\%, 4\%-9\%, and 31\%-46\% higher TBT than \sys for arXiv, Cocktail, HumanEval, and PG-19.
/AC has 6\%-9\%, 8\%-18\%, 5\%-7\%, and 11\%-19\% higher TBT than \sys.
/KE has 7\%-13\%, 10\%-21\%, 7\%-8\%, and 13\%-22\% higher TBT than \sys.
The reasons for the TBT degradation of /AC and /KE are the same as explained for TTFT. /PRC has a higher TBT degradation compared to /AC and /KE because it has $K$-decompression overhead.

\vspace{-0.1in}
\subsection{Sensitivity Testing}\label{sec:eval_sensitivity}
\vspace{-0in}

\begin{figure}[h]\vspace{-0.1in}
  \centering
  \includegraphics[width=0.5\columnwidth]{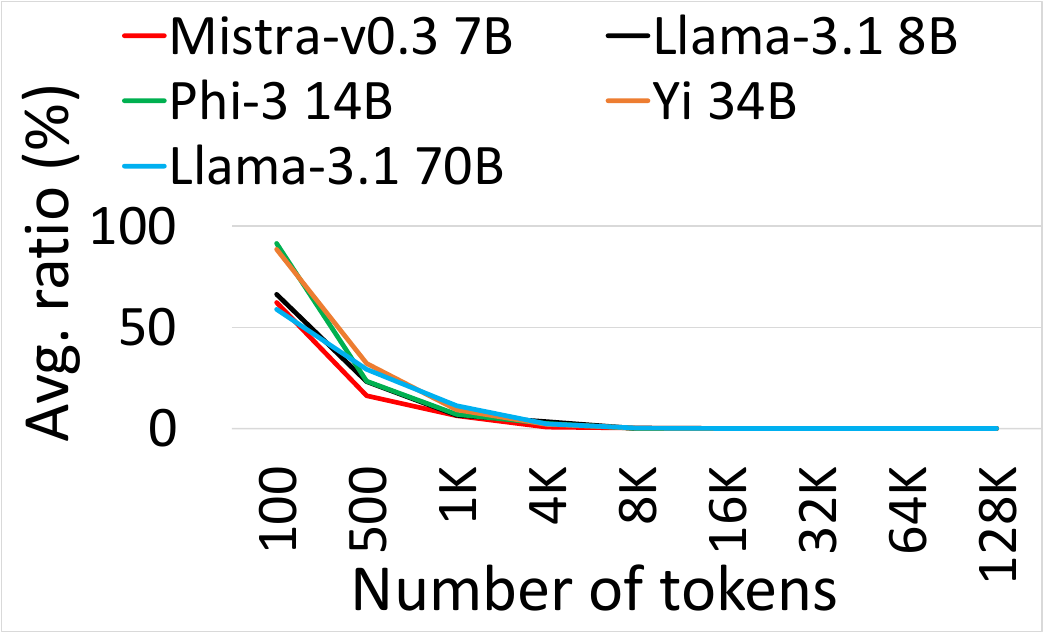}
  \vspace{-0.15in}
  \caption{The average ratio $\overline{\delta}/\overline{\varepsilon}$ across datasets.}
  \label{fig:avg_ratio_diff}
  \vspace{-0.15in}
\end{figure}

We vary the total number of random tokens used for the offline computation of $R$ to study its impact on $R$. Given a specific number of randomly generated tokens, for each model's $Q_p$-$K_p$ pair, we calculate the ratio $\overline{\delta}/\overline{\varepsilon}$ in~\cref{sec:opportunities} for each dataset and calculate their average across datasets. The smaller $\overline{\delta}/\overline{\varepsilon}$, the more accurate $R$ is. Fig.~\ref{fig:avg_ratio_diff} shows this average ratio across datasets for the $Q_p$-$K_p$ pair of each model. The results show that when the number of random tokens is greater than 8K, the average ratio $\overline{\delta}/\overline{\varepsilon}$ across datasets for all models is below 0.5\%, ensuring the accuracy of the offline-computed $R$. The rationale for this has already been explained in \cref{sec:q2} to support the conclusion in that section.

\vspace{-0.1in}
\subsection{Time Overhead}\label{sec:time_overhead}
\vspace{-0.05in}

We study the compression time overhead for multiplying $R$ online and the time overhead introduced by SM scheduling in the attention kernel. We record the number of GPU cycles occupied by the scheduling computation and use this to estimate scheduling time. Finally, we calculate the proportion of time overhead relative to JCT.
Table~\ref{tab:compress_overhead} presents the 99th percentile (P99) compression time overhead ratio for different models and datasets. The P99 compression time overhead is no more than 0.7\% of JCT for all models and datasets, indicating that the compression time overhead is negligible compared to JCT. The reason has been explained in~\cref{sec:design:decompress_attn}.
Table~\ref{tab:overhead} presents the P99 scheduling time overhead ratio for different models and datasets. As the model size increases and the input length becomes longer, the P99 time overhead ratio tends to decrease. This is because larger models and longer sequences spend more time on computation, thereby reducing the proportion of the overhead. The P99 time overhead does not exceed 4.13\% of the end-to-end latency, making the scheduling overhead acceptable.

\begin{table}[h]\vspace{-0.1in}
\small
\centering
\begin{adjustbox}{max width=\columnwidth}
\begin{tabular}{|c|c|c|c|c|}
\hline
& arXiv & Cocktail & HumanEval & PG-19 \\
\hline
Llama-3.1 8B & 0.07\% & 0.09\% & 0.3\% & 0.05\% \\
\hline
Phi-3 14B & 0.08\% & 0.04\% & 0.5\% & 0.02\% \\
\hline
Yi 34B & 0.07\% & 0.01\% & 0.7\% & 0.09\% \\
\hline
Llama-3.1 70B & 0.05\% & 0.08\% & 0.4\% & 0.04\% \\
\hline
\end{tabular}
\end{adjustbox}
\vspace{-0in}
\caption{P99 compression overhead compared to JCT.}
\vspace{-0.18in}
\label{tab:compress_overhead}
\end{table}

\begin{table}[h]\vspace{-0.1in}
\small
\centering
\begin{adjustbox}{max width=\columnwidth}
\begin{tabular}{|c|c|c|c|c|}
\hline
& arXiv & Cocktail & HumanEval & PG-19 \\
\hline
Llama-3.1 8B & 2.74\% & 2.37\% & 3.81\% & 1.27\% \\
\hline
Phi-3 14B & 2.37\% & 2.27\% & 3.54\% & 0.94\% \\
\hline
Yi 34B & 1.85\% & 1.64\% & 3.52\% & 0.85\% \\
\hline
Llama-3.1 70B & 2.39\% & 1.98\% & 3.75\% & 0.99\% \\
\hline
\end{tabular}
\end{adjustbox}
\vspace{-0in}
\caption{P99 kernel scheduling overhead compared to JCT.}
\vspace{-0.2in}
\label{tab:overhead}
\end{table}

\vspace{-0.1in}
\section{Limitations and Discussion}
\vspace{-0.0in}


\noindent\textbf{Kernel enhancement.}
The enhancement of the attention kernel with adaptive compression rates relies on a simple heuristic method for workload scheduling. In the future, we will optimize this approach to minimize attention latency. \looseness=-1

\noindent\textbf{Storage of $R$.}
The rotation matrices $R$ computed offline need to be stored in GPU memory, which incurs additional memory overhead. Fortunately, each attention head only requires a single $R$, and its shape is merely $d_h\times d_h$. The total number of parameters for all the matrices $R$ typically accounts for less than 1\% of the total parameter count of the model. Therefore, although storing $R$ introduces some memory overhead, it is negligible compared to the overall model size. \looseness=-1


\vspace{-0.1in}
\section{Related Work}
\vspace{-0.0in}

\noindent\textbf{Low-rank approximation.}
Recently, Palu~\cite{palu} has been proposed to use low-rank approximations to reduce the hidden size of KV. However, it still requires decompressing KV during attention and cannot compute on the compressed $Q$ and $K$ to reduce computation time for RoPE-based models. \sys not only eliminates decompression overhead but also reduces attention computation time by performing computation on the compressed $Q$ and $K$. 

\noindent\textbf{Eviction.}
Many KV compression methods are based on token eviction~\cite{h2o2023zhang, ge2023model, scissorhands2023liu, pyramidinfer, l2norm-kv, keyformer, dynamic-context-pruning, infinigen, zhang2024efficientsparseattentionneeds, jiang2024minference}, also known as pruning. The main idea is to remove unimportant tokens' KV that have minimal impact on inference results. The methods in~\cite{h2o2023zhang, ge2023model, scissorhands2023liu, pyramidinfer, l2norm-kv, keyformer, dynamic-context-pruning, infinigen} determine token importance based on attention scores. Tokens with lower scores are considered less important, and their KV can be discarded.
InfiniGen~\cite{infinigen} leverages low-rank approximation for QK to identify low-score tokens efficiently.
The methods in~\cite{zhang2024efficientsparseattentionneeds, jiang2024minference} achieve sparse attention by identifying low-score regions in the attention score matrix and removing KV in those regions. Eviction-based methods reduce KV along the sequence dimension, while \sys reduces KV along the hidden dimension. 

\noindent\textbf{Quantization.}
Many methods use quantization to compress KV~\cite{zipcache, kang2024gear, kivi, cachegen, kvquant}. The basic idea is to compress each element of KV by reducing high-bit representations, such as float16, to lower-bit representations, such as 4-bit or even 2-bit. It reduces the number of bits required to store KV. However, during attention computation, the quantized KV must first be dequantized to recover the original data, introducing decompression overhead. \sys, on the other hand, reduces the total number of elements of $K$ and $V$. As such, it can complement quantization methods. \sys can first compress the dimensionality of KV to reduce the total number of elements, followed by quantization to compress the individual elements.


\vspace{-0.1in}
\section{Conclusion}
\vspace{-0.0in}

In this paper, we introduce \sys, a fast KV dimensionality compression system that eliminates the $K$-decompression overhead in Palu and accelerates attention. It can serve as a complementary approach to existing eviction and quantization methods. \sys also adopts adaptive compression rates to maximize the overall KV compression rate under an accuracy constraint and leverages an enhanced attention kernel that balances SM workloads to improve attention latency.
Comprehensive experiments show that \sys reduces JCT by up to 64\% compared to Palu. When the state-of-the-art eviction and quantization methods are combined with \sys, they also outperform those combined with Palu.
\looseness=-1


\bibliographystyle{unsrt}
\bibliography{reference}

\end{document}